\documentclass[a4paper]{svproc}
\usepackage{url}
\usepackage{times}
\usepackage{algorithmic}
\usepackage{algorithm}
\usepackage{amsmath}
\usepackage{amssymb} 
\usepackage{graphics}
\usepackage{epsfig}
\usepackage[caption=false, font=footnotesize]{subfig}
\usepackage{multirow}
\usepackage{wrapfig}
\usepackage{mathtools}
\usepackage{makecell} 
\usepackage{bbm}
\usepackage{booktabs}
\usepackage{graphicx}
\usepackage{listings}
\usepackage{moresize}
\usepackage{xcolor}
\usepackage[hidelinks]{hyperref}
\usepackage{tabularx}

\newcommand{\SE}{\mathrm{SE}}

\begin{document}
\mainmatter

\title{BulletArm: An Open-Source Robotic Manipulation Benchmark and Learning Framework}
\titlerunning{BulletArm} 
\author{Dian Wang\textsuperscript{*} \and 
        Colin Kohler\textsuperscript{*} \and 
        Xupeng Zhu \and Mingxi Jia \and 
        Robert Platt}
\authorrunning{Wang et al.} 
\institute{Khoury College of Computer Sciences\\
Northeastern University\\
Boston, MA 02115, USA\\
\email{\{wang.dian, kohler.c, zhu.xup, jia.ming, r.platt\}@northeastern.edu}
\footnote[0]{\textsuperscript{*}Equal Contribution}
}

\maketitle

\begin{abstract}
We present BulletArm, a novel benchmark and learning-environment for robotic manipulation.
BulletArm is designed around two key principles: reproducibility and extensibility. We aim
to encourage more direct comparisons between robotic learning methods by providing a set
of standardized benchmark tasks in simulation alongside a collection of baseline algorithms.
The framework consists of 31 different manipulation tasks of varying difficulty, 
ranging from simple reaching and picking tasks to more realistic tasks such as bin 
packing and pallet stacking. In addition to the provided tasks, BulletArm has been built
to facilitate easy expansion and provides a suite of tools to assist users when 
adding new tasks to the framework.
Moreover, we introduce a set of five benchmarks and evaluate them using a series of
state-of-the-art baseline algorithms. By including these algorithms as part of our 
framework, we hope to encourage users to benchmark their work on any new tasks against
these baselines.

\keywords{Benchmark, Simulation, Robotic Learning, Reinforcement Learning}
\end{abstract}
 
\section{Introduction}

Inspired by the recent successes of deep learning in the field of computer vision,
there has been an explosion of work aimed at applying deep learning 
algorithms across a variety of disciplines. Deep reinforcement learning, for example, 
has been used to learn policies which achieve superhuman levels of performance across 
a variety of games \cite{alphago,dqn}. Robotics has seen a similar surge
in recent years, especially in the area of robotic manipulation with reinforcement
learning \cite{gu2017deep,qt_opt,zeng_push}, imitation learning 
\cite{transporter}, and multi-task learning \cite{devin2017learning,hausman2018learning}.
However, there is a key difference between current robotics learning research and
past work applying deep learning to other fields. There currently is no widely 
accepted standard for comparing learning-based robotic manipulation methods.
In computer vision for example, the ImageNet benchmark \cite{deng2009imagenet} has
been a crucial factor in the explosion of image classification algorithms we have 
seen in the recent past.

While there are benchmarks for policy learning in domains similar to robotic
manipulation, such as the continuous control tasks in OpenAI Gym \cite{gym} and
the DeepMind Control Suite~\cite{dmc}, they are not applicable to more real-world
tasks we are interested in robotics. Furthermore, different robotics labs work with 
drastically different systems in the real-world using different robots, 
sensors, etc. As a result, researchers often develop their own training and evaluation 
environments, making it extremely difficult to compare different approaches. 
For example, even simple tasks like block stacking can have a lot of variability
between different works \cite{nair2018overcoming,li2020towards,asr},
including different physics simulators, different manipulators, different object sizes, etc.

In this work, we introduce BulletArm, a novel framework for robotic manipulation 
learning based on two key components. First, we provide a flexible, open-source framework
that supports many different manipulation tasks. Compared with prior works, we introduce 
tasks with various difficulties and require different manipulation skills. This includes 
long-term planning tasks like supervising Covid tests and contact-rich tasks requiring 
precise nudging or pushing behaviors. BulletArm currently consists of 31 unique tasks 
which the user can easily customize to mimic their real-world lab setups (e.g., workspace 
size, robotic arm type, etc). In addition, BulletArm was developed with a emphasis on 
extensability so new tasks can easily be created as needed. Second, we include five different
benchmarks alongside a collection of standardized baselines for the user to quickly benchmark
their work against. We include our implementations of these baselines in the hopes of new 
users applying them to their customization of existing tasks and whatever new
tasks they create.

Our contribution can be summarized as three-fold. First, we propose BulletArm, a 
benchmark and learning framework containing a set of 21 open-loop manipulation tasks
and 10 close-loop manipulation tasks. We have developed this framework over the course
of many prior works \cite{asr,ondrej_action_prior,equi_asr,ondrej_action_prior,equi_rl,equi_sac_robot,equi_grasp}.
Second, we provide state-of-the-art baseline algorithms enabling other researchers to
easily compare their work against our baselines once new tasks are inevitably added to the
baseline.
Third, BulletArm provides a extensive suite of tools to allow users to easily create
new tasks as needed. Our code is available at \url{https://github.com/ColinKohler/BulletArm}. 

\section{Related Work}
\paragraph{Reinforcement Learning Environments}
Standardized environments are vitally important when comparing different reinforcement 
learning algorithms. Many prior works have developed various video game environments, 
including PacMan~\cite{pacman}, Super Mario \cite{mario}, Doom~\cite{vizdoom}, and
StarCraft~\cite{starcraft}. OpenAI Gym~\cite{gym} provides a standard API for the 
communication between agents and environments, and a collection of various different
environments including Atari games from the Arcade Learning Environment (ALE)~\cite{atari}
and some robotic tasks implemented using MuJoCo~\cite{mujoco}. The DeepMind 
Control Suite (DMC)~\cite{dmc} provides a similar set of continuous control tasks. 
Although both OpenAI Gym and DMC have a small set of robotic environments, 
they are toy tasks which are not representative of the real-world tasks we are 
interested in robotics.

\noindent
\paragraph{Robotic Manipulation Environments}
In robotic manipulation, there are many benchmarks for grasping in the 
context of supervised learning, e.g., the Cornell dataset~\cite{cornell}, the 
Jacquard dataset~\cite{jacquard}, and the GraspNet 1B dataset~\cite{graspnet_1b}. 
In the context of reinforcement learning, on the other hand, the majority of prior 
frameworks focus on single tasks, for example, door opening~\cite{door_gym}, 
furniture assembly~\cite{ikea_task}, and in-hand dexterous
manipulation~\cite{learning_dexterous}. Another strand of prior works propose 
frameworks containing a variety of different environments, such as 
robosuite~\cite{robosuite}, PyRoboLearn~\cite{pyrobolearn}, and
Meta-World~\cite{meta_world}, but are often limited to short horizon tasks.
Ravens~\cite{transporter} introduces a set of environments containing complex 
manipulation tasks but restricts the end-effector to a suction cup gripper.
RLBench~\cite{rlbench} provides a similar learning framework to ours with a number 
of key differences. First, RLBench is built around the PyRep \cite{james2019pyrep} 
interface and is therefor built on-top of V-REP \cite{rohmer2013vrep}.
Furthermore, RLBench is more restrictive than BulletArm with limitations placed 
on the workspace scene, robot, and more. 

\noindent
\paragraph{Robotic Manipulation Control}
There are two commonly used end-effector control schemes: open-loop control and 
close-loop control. In open-loop control, the agent selects both the target pose of
target pose of the end-effector and some action primitive to execute at that pose.
Open-loop control generally has shorter time horizon, allowing the agent to solve 
complex tasks that require a long trajectory \cite{zeng_pick,zeng_push,asr}.
In close-loop control, the agent sets the displacement of the end-effector This
allows the agent to more easily recover from failures which is vital when delaing
with contact-rich tasks \cite{gu2017deep,qt_opt,ferm,equi_rl}.
BulletArm provides a collection of environments in both settings, allowing the users
to select either one based on their research interests.

\section{Architecture}
\label{sec:architecture}
At the core of our learning framework is the PyBullet \cite{pybullet} simulator. 
PyBullet is a Python library for robotics simulation and machine learning with a
focus on sim-to-real transfer. Built upon Bullet Physics SDK, PyBullet provides 
access to forward dynamics simulation, inverse dynamics computation, forward and
inverse kinematics, collision detection, and more. In addition to physics simulation,
there are also numerous tools for scene rendering and visualization. BulletArm
builds upon PyBullet, providing a diverse set of tools tailored to robotic 
manipulation simulations.

\subsection{Design Philosophy}
The design philosophy behind our framework focuses on four key principles:

\textbf{1. Reproducibility:} 
A key challenge when developing new learning algorithms is the difficulty
in comparing them to previous work. In robotics, this problem is especially 
prevalent as different researchers have drastically different robotic setups. 
This can range from small differences, such as workspace size or degradation of
objects, to large differences such as the robot used to preform the experiments.
Moving to simulation allows for the standardization of these factors but can 
impact the performance of the trained algorithm in the real-world. We aim
to encourage more direct comparisons between works by providing a flexible 
simulation environment and a number of baselines to compare against. 

\textbf{2. Extensibility:} 
Although we include a number of tasks, control types, and robots; there will
always be a need for additional development in these areas. Using our framework,
users can easily add new tasks, robots, and objects. We make the choice to not 
restrict tasks, allowing users more freedom create interesting domains. 
Figure~\ref{fig:api_example} shows an example of creating a new task 
using our framework.

\textbf{3. Performance:} 
Deep learning methods are often time consuming, slow processes and the addition
of a physics simulator can lead to long training times. We have spent a significant
portion of time in ensuring that our framework will not bottleneck training by
optimizing the simulations and allowing the user to run many environments in 
parallel. 

\textbf{4. Usability:} 
A good open-source framework should be easy to use and understand. We provide
extensive documentation detailing the key components of our framework and a 
set of tutorials demonstrating both how to use the environments and how to
extend them.

\subsection{Environment}

\begin{figure}[t]
\centering
\setlength{\tabcolsep}{1.0em} 
\begin{tabular}{cc}
\begin{lstlisting}[
language=Python,
basicstyle={\ssmall\ttfamily},
keywordstyle=\color{magenta},
tabsize=2,
numbers=left,
]
from bulletarm import env_factory

task_config = {'robot': 'kuka'}
env = env_factory.createEnvs(1, 
    'block_stacking', task_config)
agent = Agent()
obs = env.reset()
while not done:
  if expert:
    action = env.getNextAction()
  else:
    action = agent.getAction(obs)
  obs, reward, done = env.step(action)
env.close()
\end{lstlisting}
& 
\begin{lstlisting}[
language=Python,
basicstyle={\ssmall\ttfamily},
keywordstyle=\color{magenta},
tabsize=2,
numbers=left,
]
from bulletarm.base_env import BaseEnv 
from bulletarm.constants import CUBE

class PyramidStackEnv(BaseEnv):
  def __init__(self, config):
    super().__init__(config)
    
  def reset(self):
    self.resetPybulletWorkspace()
    self.cubes = self._generateShapes(CUBE, 3)
    return self._getObservation()
    
  def _checkTermination(self):
    return self.areBlocksInPyramid(self.cubes) 
\end{lstlisting}
\\
\end{tabular}
\caption{Example scripts using our framework. (Left) Creating and interacting
         with a environment running the Block Stacking task. (Right) Creating a new block 
         structure construction task by subclassing the existing base domain.}
\label{fig:api_example}
\end{figure}

\begin{figure}[t]
\centering
\subfloat[Kuka]{
\includegraphics[width=0.2\linewidth]{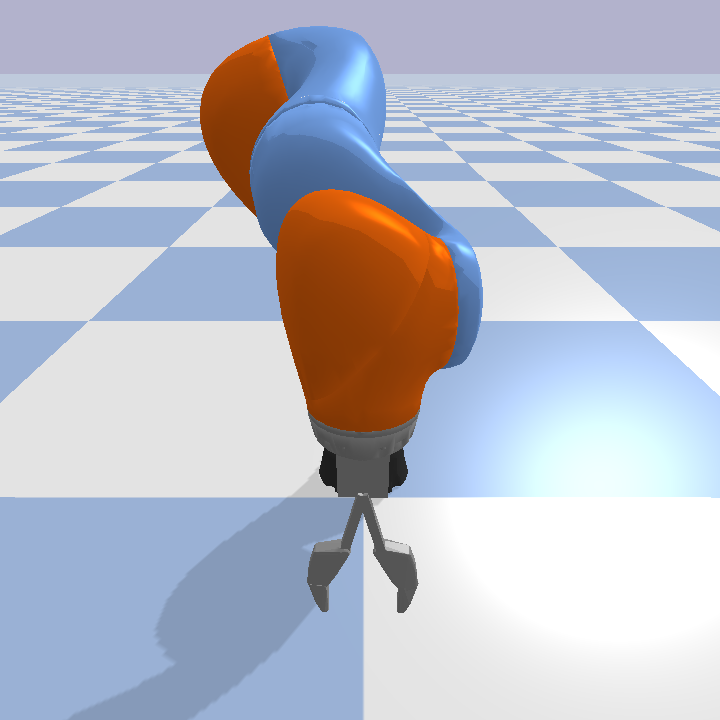}
}
\subfloat[Panda]{
\includegraphics[width=0.2\linewidth]{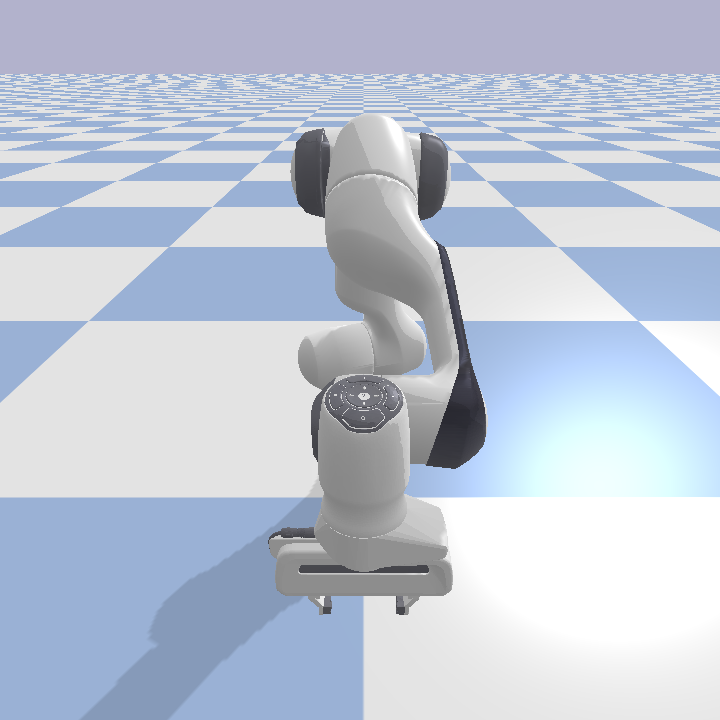}
}
\subfloat[UR5 Parallel]{
\includegraphics[width=0.2\linewidth]{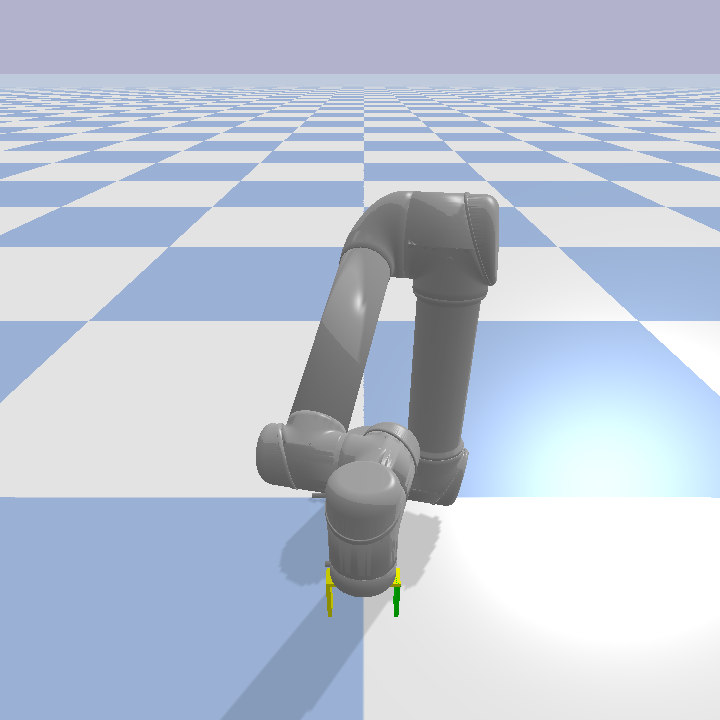}
}
\subfloat[UR5 Robotiq]{
\includegraphics[width=0.2\linewidth]{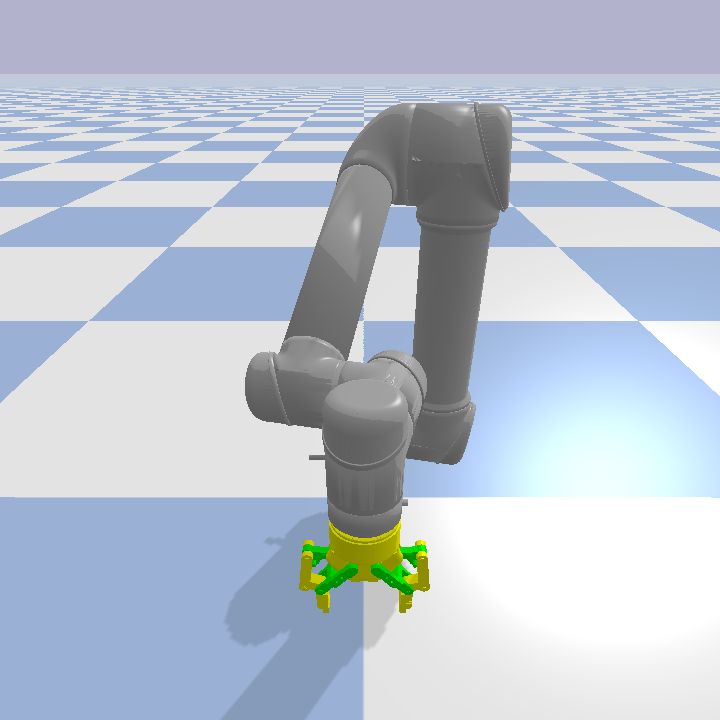}
}
\caption{Our work currently supports four different arms: Kuka, Panda, UR5 with 
         parallel jaw gripper, and UR5 with Robotiq gripper.}
\label{fig:robot}
\end{figure}

Our simulation setup (Figure \ref{fig:open_loop_workspace}) consists of a robot 
arm mounted on the floor of the simulation environment, a workspace in front of 
the robot where objects are generated, and a sensor. Typically, we use top-down
sensors which generate heightmaps of the workspace. As we restrict the perception 
data to only the defined workspace, we choose to not add unnecessary elements to
the setup such as a table for the arm to sit upon. Currently there are four different 
robot arms available in BulletArm (Figure \ref{fig:robot}): KUKA IIWA, Frane Emika Panda, 
Universal Robots UR5 with either a simple parallel jaw gripper or the Robotiq 2F-85 
gripper. 

\textit{Environment}, \textit{Configuration}, and \textit{Episode} are three key terms 
within our framework. An environment is an instance of the PyBullet simulator 
in which the robot interacts with objects while trying to solve some task. This 
includes the initial environment state, the reward function, and termination 
requirements. A configuration contains additional specifications for the task
such as the robotic arm, the size of the workspace, the physics mode, etc (see
Appendix~\ref{appendix:parameters} for an full list of parameters).
Episodes are generated by taking actions (steps) within an environment until the 
episode ends. An episode trajectory $\tau$ contains a series of observations $o$,
actions $a$, and rewards $r$: $\tau = [(o_0,a_0,r_0), ..., (o_T,a_T,r_T)]$.

Users interface with the learning environment through the 
\textit{EnvironmentFactory} and \textit{EnvironmentRunner} classes. The 
EnvironmentFactory is the entry point and creates the \textit{Environment} class
specified by the Configuration passed as input. The EnvironmentFactory can create 
either a single environment or multiple environmaents meant to 
be run in parallel. In either case, an EnvironmentRunner instance is returned and
provides the API which interacts with the environments. This API,
Figure \ref{fig:api_example}, is modelled after the typical agent-environment 
RL setup popularized by OpenAI Gym \cite{gym}. 

The benchmark tasks we provide have a sparse reward function which returns $+1$
on successful task completion and $0$ otherwise. While we find this reward function
to be advantageous as it avoids problems due to reward shaping, we do not require 
that new tasks conform to this. When defining a new task, the reward
function defaults to sparse but users can easily define their custom reward
for a new task. We separate our tasks into two categories based on the action 
spaces: open-loop control and closed-loop control. These two control modes are 
commonly used in robotics manipulation research.

\subsection{Expert Demonstrations}
\begin{figure}[t]
\centering
\includegraphics[width=0.9\linewidth]{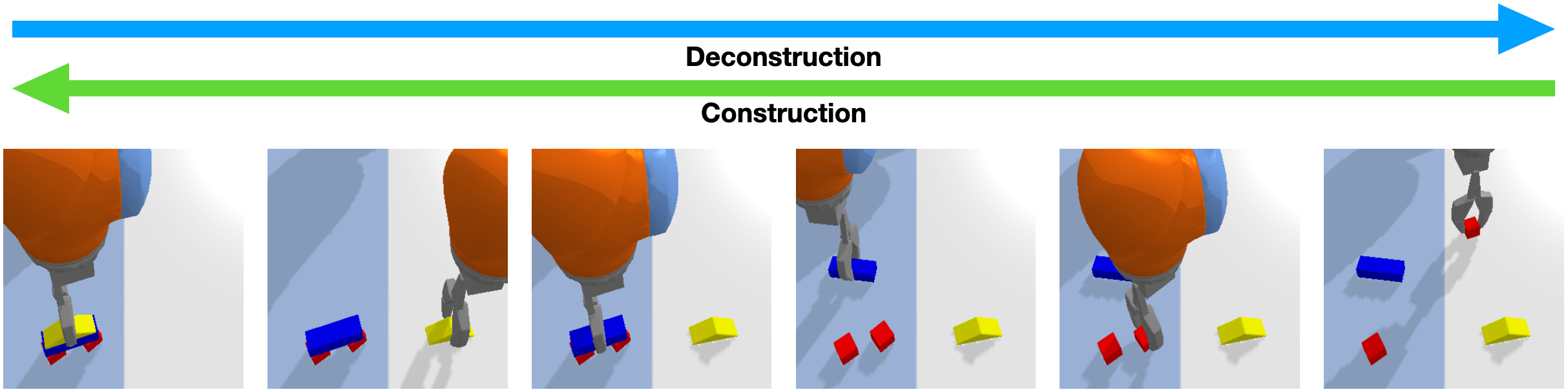}
\caption{The Deconstruction planner. Left to right: a deconstruction episode 
         where the expert deconstructs the block structure in the left-most 
         figure. Right to left: a construction episode is generated by reversing 
         the deconstruction episode. This is inspired by~\cite{form2fit} where 
         the authors propose a method to learn kit assembly through disassembly
         by reversing disassembly transitions.}
\label{fig:decon_planner}
\end{figure}

Expert demonstrations are crucial to many robotic learning fields. Methods 
such as imitation and model-based learning, for example, learn directly from
expert demonstrations. Additionally, we find that in the context of reinforcement
learning, it is vital to seed learning with expert demonstrations due to 
the difficulties in exploring large state-action spaces. We provide two types 
of planners to facilitate expert data generation: the 
\textit{Waypoint Planner} and the \textit{Deconstruction Planner}.

The Waypoint Planner is a online planning method which moves the end-effector through
a series of waypoints in the workspace. We define a waypoint as $w_t = (p_t, a_t)$ where
$p_t$ is the desired pose of the end effector and $a_t$ is the action primitive to execute
at that pose. These waypoints can either be absolute positions in the workspace or positions
relative to the objects in the workspace. In open-loop control, the planner returns the
waypoint $w_t$ as the action executed at time $t$. In close-loop control, the planner will
continuously return a small displacement from the current end-effector pose as the action at 
time $t$. This process is repeated until the waypoint has been reached.
The Deconstruction Planner is a more limited planning method which can only be applied 
to pick-and-place tasks where the goal is to arrange objects in a specific manner. For 
example, we utilize this planner for the various block construction tasks examined in 
this work (Figure \ref{fig:decon_planner}). When using this planner, the workspace is
initialized with the objects in their target configuration and objects are then removed
one-by-one until the initial state of the task is reached. This deconstruction trajectory,
is then reversed to produce an expert construction trajectory, 
$\tau_{expert} = \textit{reverse}(\tau_{deconstruct})$.

\section{Environments}

\begin{figure}[t]
\begin{minipage}{.39\textwidth}
\centering
\includegraphics[width=0.85\linewidth]{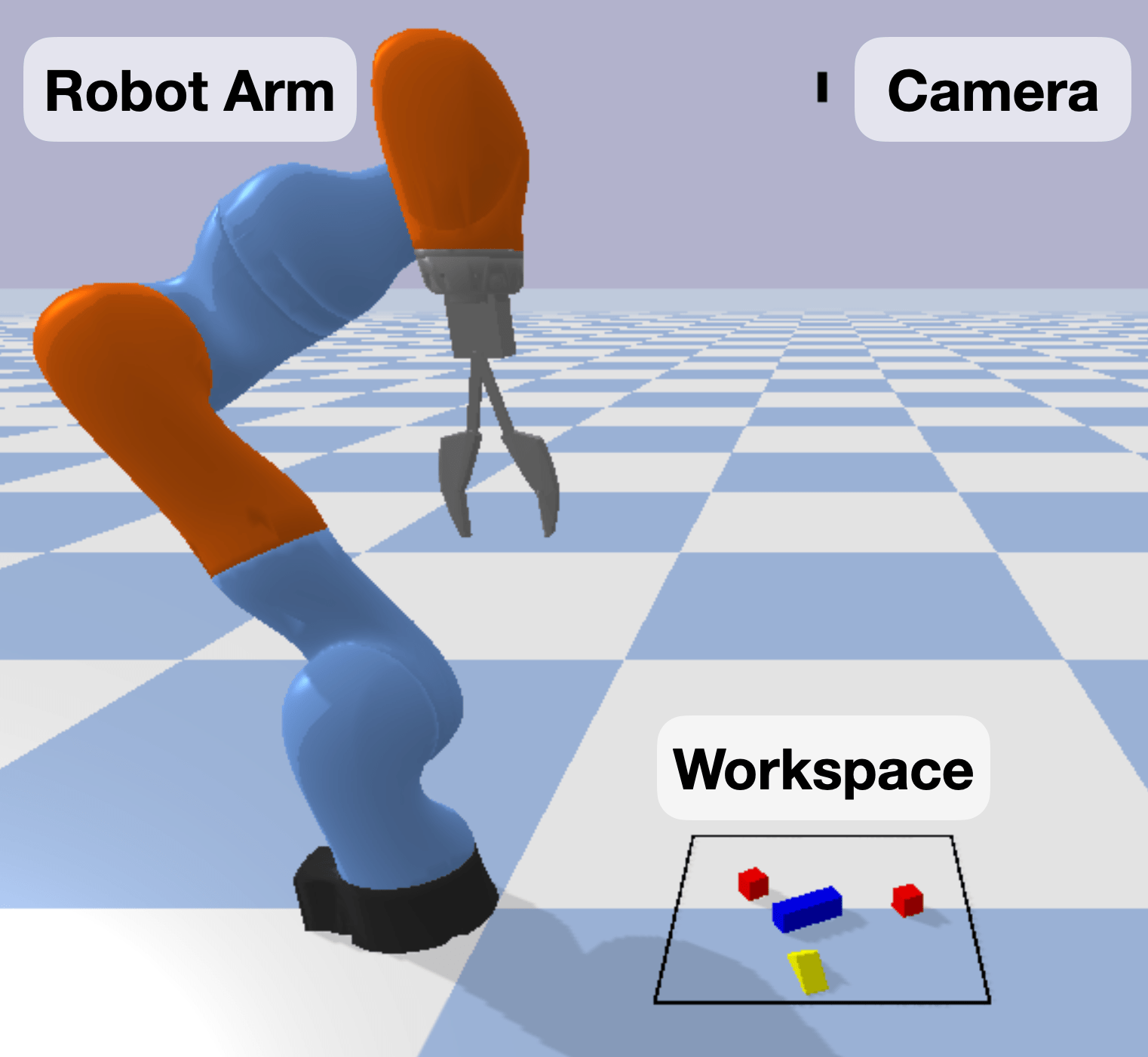}
\caption{The environment containing a robot arm, a camera, and a workspace}
\label{fig:open_loop_workspace}
\end{minipage}
\hspace{0.1cm}
\begin{minipage}{.59\textwidth}
\centering
\subfloat[]{
\includegraphics[width=0.34\linewidth]{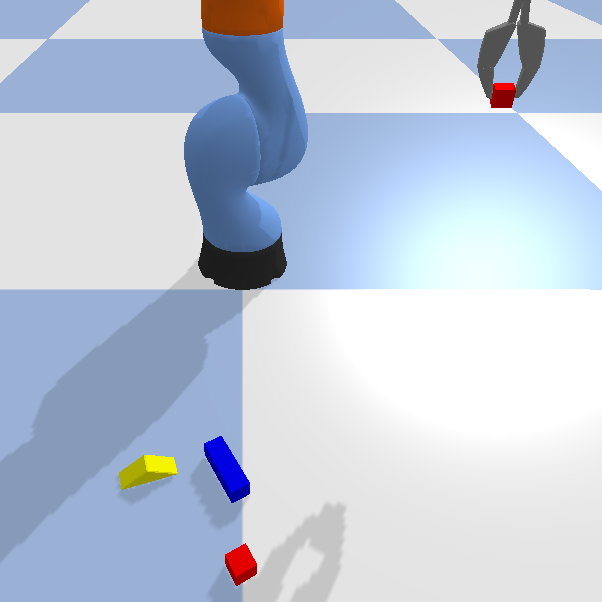}
}
\subfloat[]{
\includegraphics[width=0.61\linewidth]{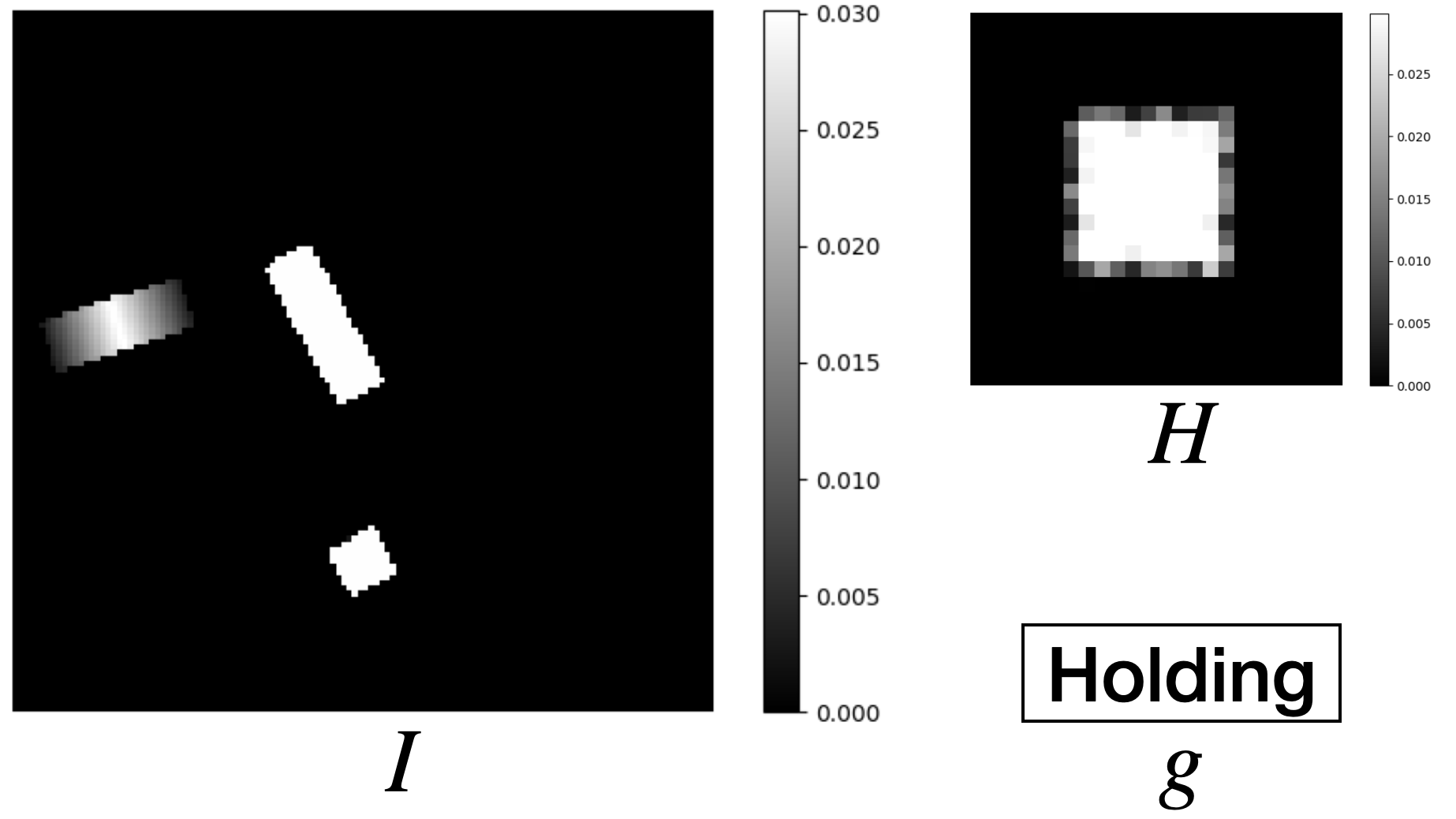}
}
\caption{(a) The manipulation scene. (b) The state including a top-down heightmap $I$, an in-hand image $H$ and the gripper state $g$.}
\label{fig:open_loop_obs}
\end{minipage}
\end{figure}

The core of any good benchmark is its set of environments. In robotic 
manipulation, in particular, it is important to cover a broad range of task
difficulty and diversity. To this end, we introduce tasks covering a variety of
skills for both open-loop and close-loop control. Moreover, the configurable 
parameters of our environments enable the user to select different task variations
(e.g., the user can select whether the objects in the workspace will be initialized
with a random orientation). BulletArm currently provides a collection of 21 
open-loop manipulation environments and a collection of 10 close-loop environments.
These environments are limited to kinematic tasks where the robot has to directly 
manipulate a collection of objects in order to reach some desired configuration.

\subsection{Open-Loop Environments}
\label{sec:open_envs}

\begin{figure}[t]
\captionsetup[subfloat]{format=hang,justification=centering}
\newlength{\env}
\setlength{\env}{0.23\linewidth}
\centering
\subfloat[Block Stacking]{
\includegraphics[width=\env]{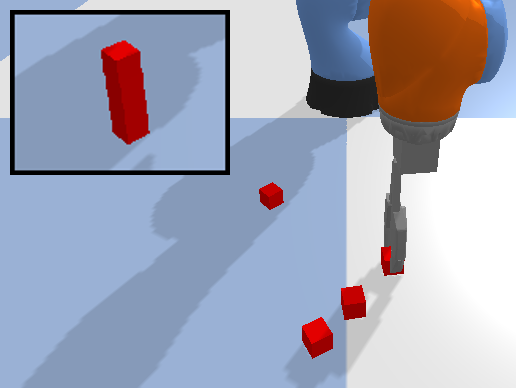}
\label{fig:open_stack}
}
\subfloat[House Building 1]{
\includegraphics[width=\env]{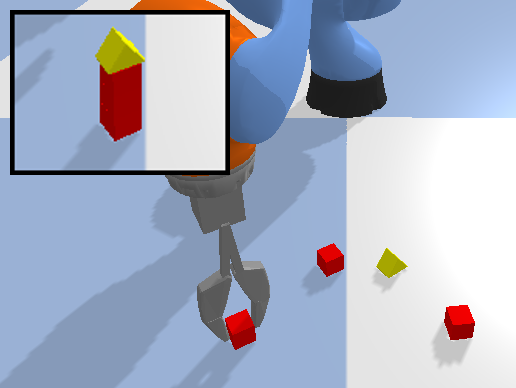}
\label{fig:open_h1}
}
\subfloat[House Building 2]{
\includegraphics[width=\env]{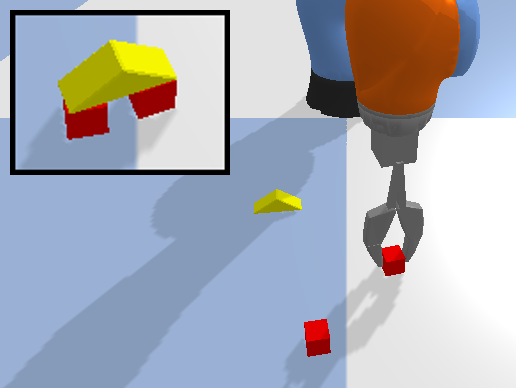}
\label{fig:open_h2}
}
\subfloat[House Building 3]{
\includegraphics[width=\env]{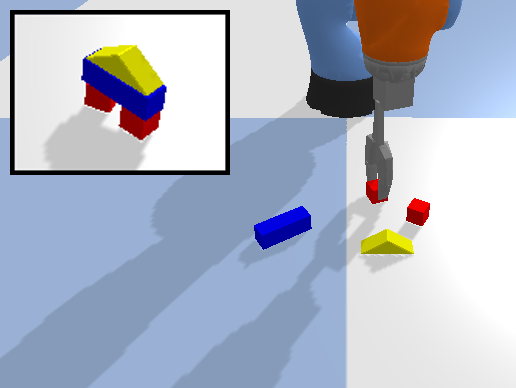}
\label{fig:open_h3}
}\\
\subfloat[House Building 4]{
\includegraphics[width=\env]{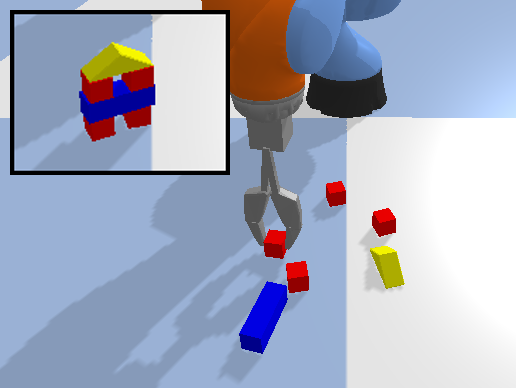}
\label{fig:open_h4}
}
\subfloat[Improvise House\\ Building 2]{
\includegraphics[width=\env]{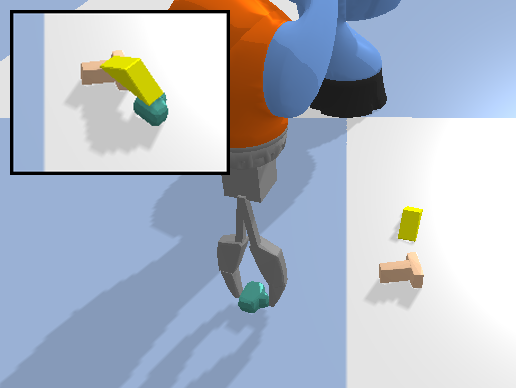}
\label{fig:open_imh2}
}
\subfloat[Improvise House\\ Building 3]{
\includegraphics[width=\env]{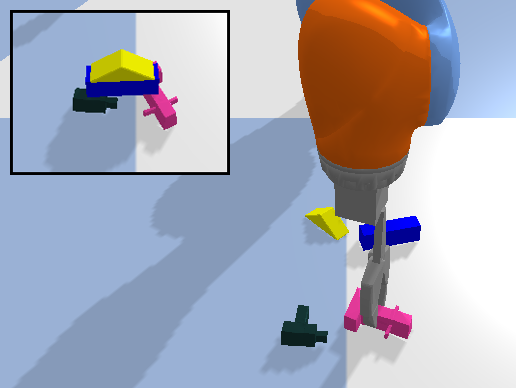}
\label{fig:open_imh3}
}
\subfloat[Bin Packing]{
\includegraphics[width=\env]{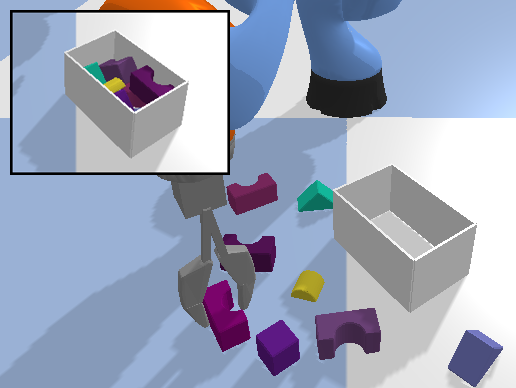}
\label{fig:open_pack}
}\\
\subfloat[Bottle Arrangement]{
\includegraphics[width=\env]{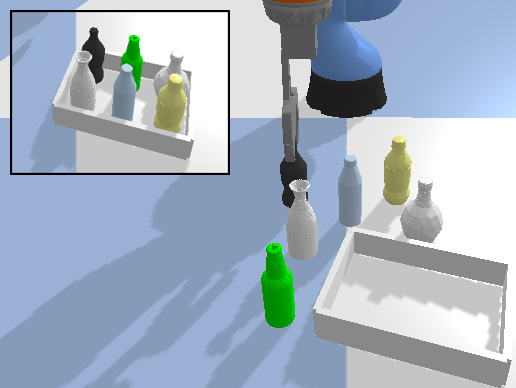}
\label{fig:open_bottle}
}
\subfloat[Box Palletizing]{
\includegraphics[width=\env]{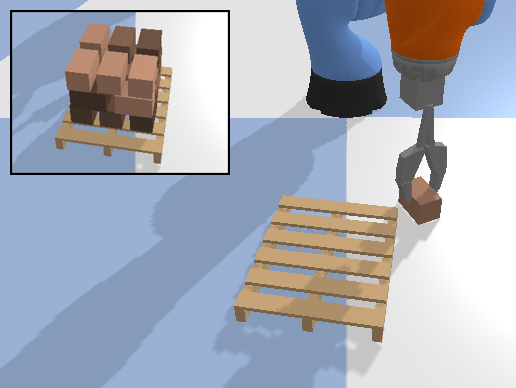}
\label{fig:open_box}
}
\subfloat[Covid Test]{
\includegraphics[width=\env]{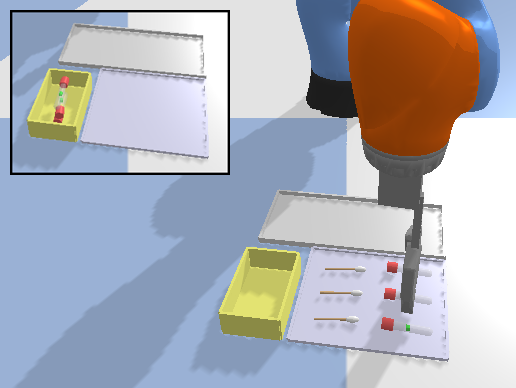}
\label{fig:open_covid}
}
\subfloat[Object Grasping]{
\includegraphics[width=\env]{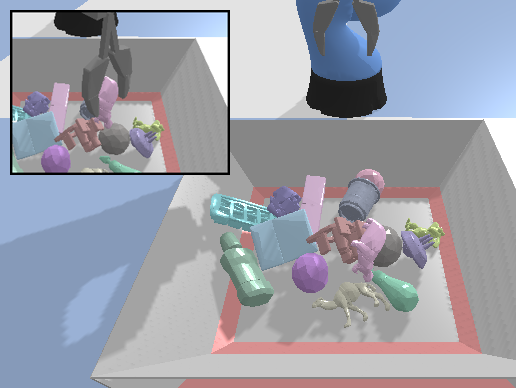}
\label{fig:open_grasp}
}
\caption{The open-loop environments. The window on the top-left corner of each sub-figure shows the goal state of each task.}
\label{fig:open_loop_envs}
\end{figure}

In the open-loop environments, the agent controls the target pose of the end-effector,
resulting in a shorter time horizon for complex tasks. The open-loop environment is 
comprised of a robot arm, a workspace, and a camera above the workspace providing 
the observation (Figure~\ref{fig:open_loop_workspace}). The action space is defined 
as the cross product of the gripper motion $A_g=\{\textsc{pick}, \textsc{place}\}$ and
the target pose of the gripper for that motion $A_p$, $A=A_g\times A_p$. The state space is 
defined as $s=(I, H, g)\in S$ (Figure~\ref{fig:open_loop_obs}), where $I$ is a 
top-down heightmap of the workspace; $g\in \{\textsc{holding}, \textsc{empty}\}$ 
denotes the gripper state; and $H$ is an in-hand image that shows the object 
currently being held. If the last action is $\textsc{pick}$, then $H$ is a crop of 
the heightmap in the previous time step centered at the pick position. If the last
action is $\textsc{place}$, $H$ is set to a zero value image.

BulletArm provides three different action spaces for $A_p$: 
$A_p\in \{A^{xy}, A^{xy\theta}, A^{\SE(3)}\}$.
The first option $(x, y)\in A^{xy}$ only controls the $(x, y)$ components of the 
gripper pose, where the rotation of the gripper is fixed; and $z$ is selected using 
a heuristic function that first reads the maximal height in the region around
$(x, y)$ and then either adds a positive offset for a $\textsc{place}$ action or a 
negative offset for $\textsc{pick}$ action. The second option 
$(x, y, \theta)\in A^{xy\theta}$ adds control of the rotation $\theta$ 
along the $z$-axis. The third option $(x, y, z, \theta, \phi, \psi)\in A^{\SE(3)}$ 
controls the full 6 degree of freedom pose of the gripper, including the rotation along 
the $y$-axis $\phi$ and the rotation along the $x$-axis $\psi$. $A^{xy}$ and $A^{xy\theta}$ 
are suited for tasks that only require top-down manipulations, while $A^{\SE(3)}$ is 
designed for solving complex tasks involving out-of-plane rotations. The definition 
of the state space and the action space in the open-loop environments also enables 
effortless sim2real transfer. One can reproduce the observation in Figure~\ref{fig:open_loop_obs} 
in the real-world using an overhead depth camera and transfer the learned policy~\cite{asr,equi_asr}.

Figure~\ref{fig:open_loop_envs} shows the 12 basic open-loop environments that can be solved using top-down actions. Those 
environments can be categorized into two collections, a set of block structure 
tasks (Figures~\ref{fig:open_stack}-\ref{fig:open_imh3}), and a 
set of more realistic tasks (Figures \ref{fig:open_pack} and 
\ref{fig:open_grasp}). The block structure tasks require the robot to build 
different goal structures using blocks. The more realistic tasks require 
the robot to finish some real-world problems, for example, arranging bottles or 
supervising Covid tests. We use the default sparse reward function for
all open-loop environments, i.e., +1 reward for reaching the goal, and 0 otherwise. 
See Appendix~\ref{appendix:open_envs} for a detailed description of the tasks. 

\begin{figure}[t]
\centering
\subfloat[The Ramp Environment]{
\includegraphics[width=0.2\linewidth]{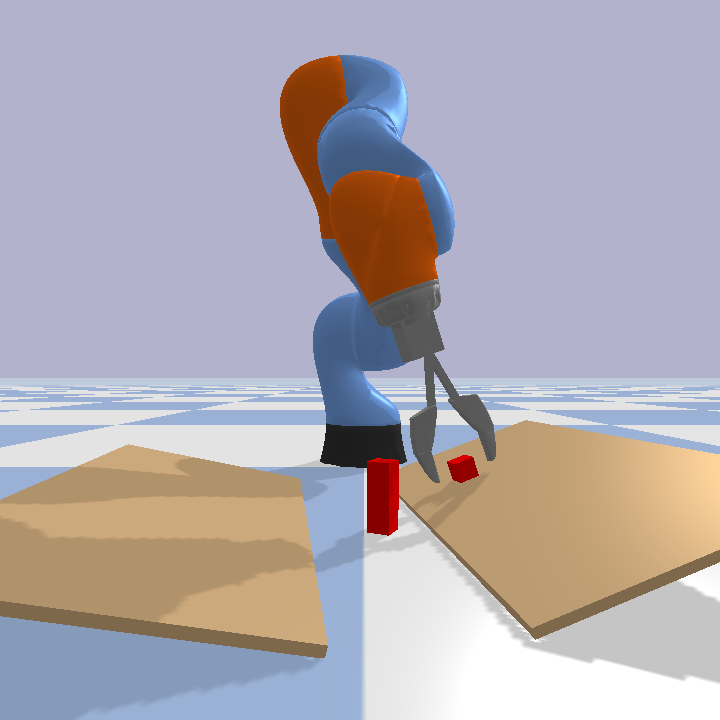}
\includegraphics[width=0.2\linewidth]{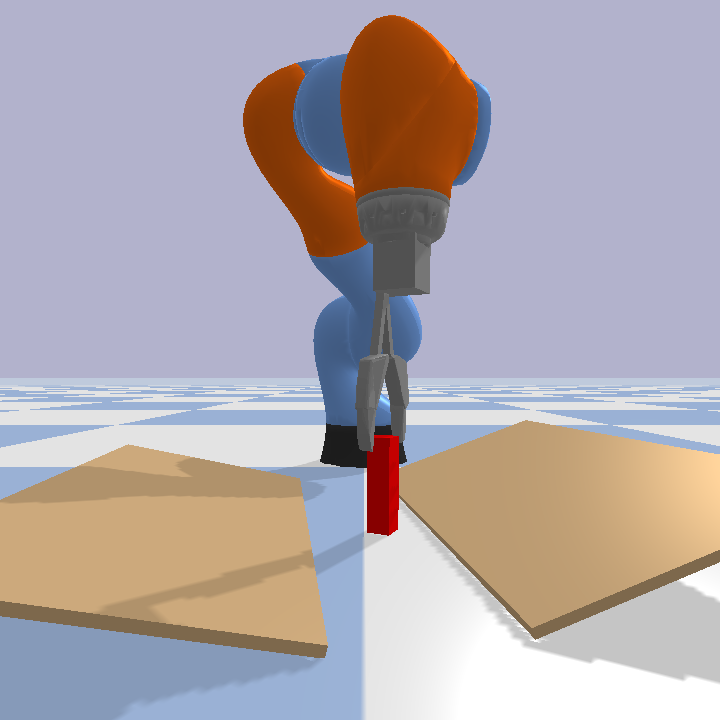}
}
\subfloat[The Bump Environment]{
\includegraphics[width=0.2\linewidth]{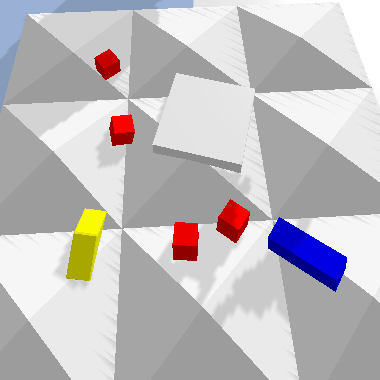}
\includegraphics[width=0.2\linewidth]{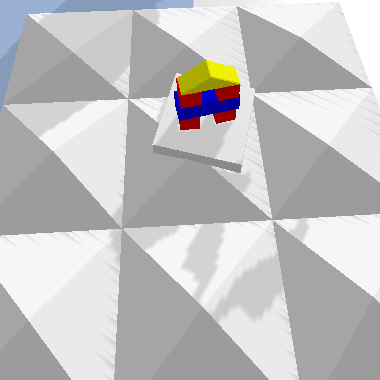}
}
\caption{The 6DoF environments. (a): In the Ramp Environment, the objects are initialized on two ramps, where the agent needs to control the out-of-plane orientations to pick up the objects. (b): Similarly, in the Bump Environment, the objects are initialized on a bumpy surface.}
\label{fig:open_6d_env}
\end{figure}

\begin{figure}[t]
\centering
\subfloat[]{
\includegraphics[height=0.26\linewidth]{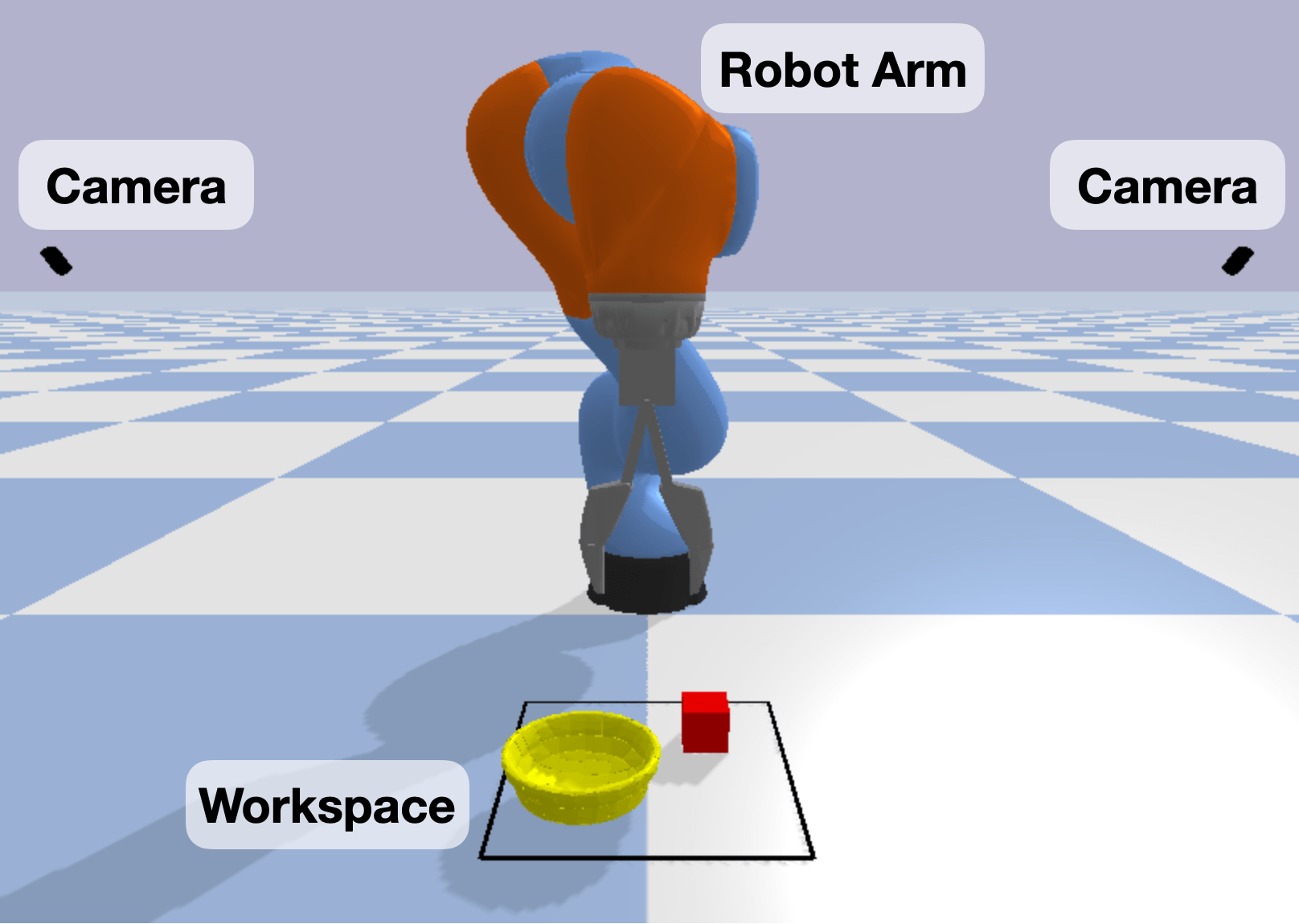}
\label{fig:close_ws}
}
\subfloat[]{
\includegraphics[height=0.26\linewidth]{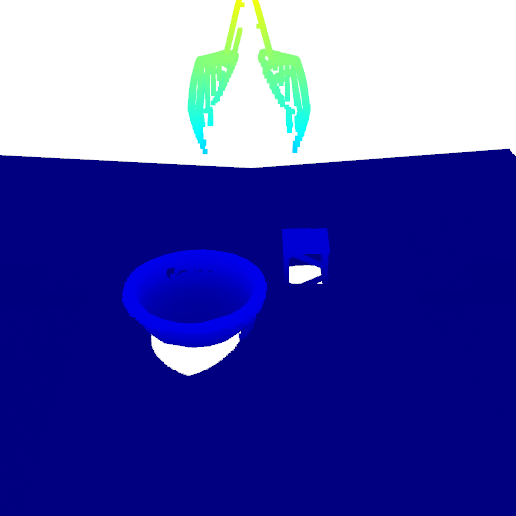}
\label{fig:close_pc}
}
\subfloat[]{
\includegraphics[height=0.26\linewidth]{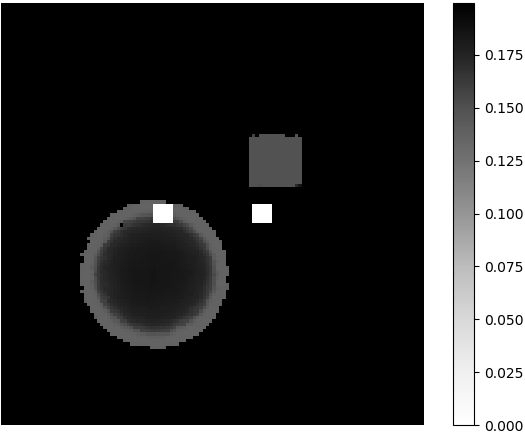}
\label{fig:close_obs}
}
\caption{(a) The close-loop environment containing a robot arm, two cameras, and a workspace. (b) The point cloud generated from the two cameras. (c) The orthographic projection generated from the point cloud which is used as the observation. The two squares at the center of the image represent the gripper. Alternatively, the image can be generated using a simulated orthographic camera located at the position of the end-effector.}
\end{figure}

\subsubsection{6DoF Extensions} The environments that we have introduced so far 
only require the robot to perform top-down manipulation. We extend those environments
to 6 degrees of freedom by initializing them in either the ramp environment or the bump 
environment (Figure~\ref{fig:open_6d_env}). In both cases, the robot needs to control
the out-of-plane orientations introduced by the ramp or bump in order to manipulate
the objects. We provide seven ramp environments (for each of the block structure
construction tasks), and two bump environments (House Building 4 and Box Palletizing). 
See Appendix~\ref{appendix:6dof} for details.

\begin{figure}[t]
\setlength{\env}{0.23\linewidth}
\centering
\subfloat[Block Reaching]{
\includegraphics[width=\env]{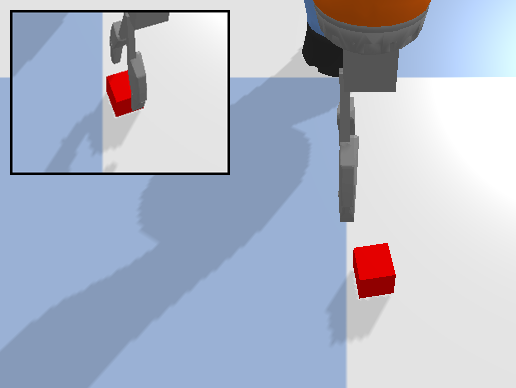}
\label{fig:close_reach}
}
\subfloat[Block Picking]{
\includegraphics[width=\env]{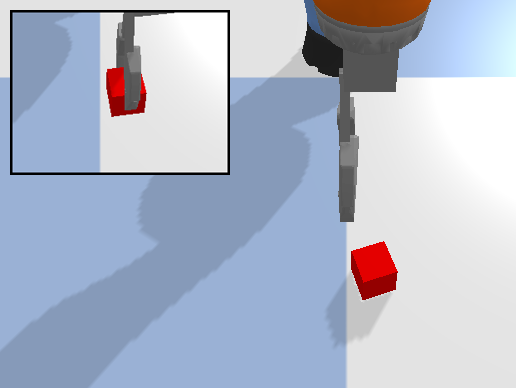}
\label{fig:close_pick}
}
\subfloat[Block Pushing]{
\includegraphics[width=\env]{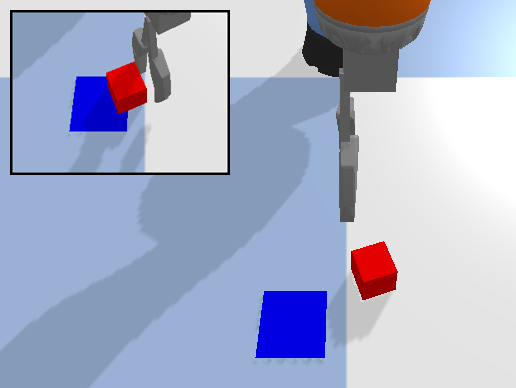}
\label{fig:close_push}
}
\subfloat[Block Pulling]{
\includegraphics[width=\env]{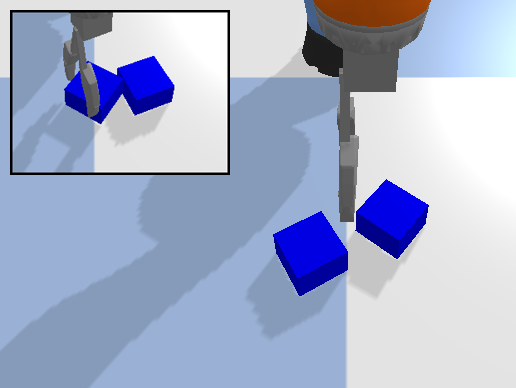}
\label{fig:close_pull}
}\\
\subfloat[Block in Bowl]{
\includegraphics[width=\env]{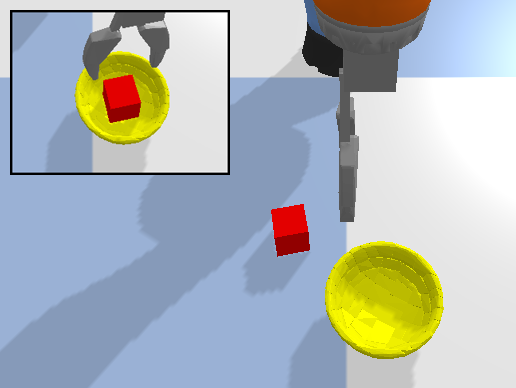}
\label{fig:close_bowl}
}
\subfloat[Block Stacking]{
\includegraphics[width=\env]{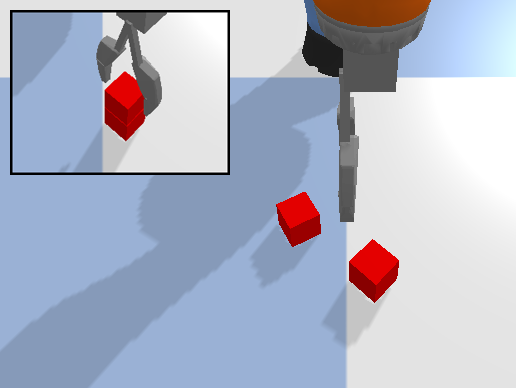}
\label{fig:close_stack}
}
\subfloat[House Building]{
\includegraphics[width=\env]{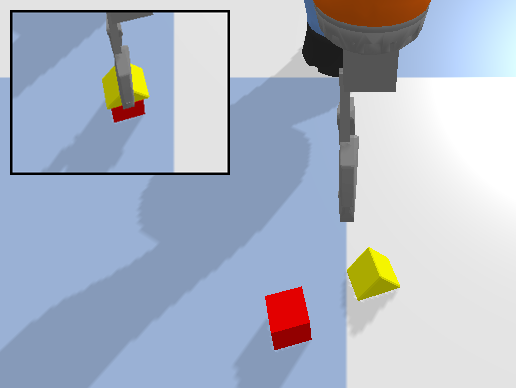}
\label{fig:close_h1}
}
\subfloat[Corner Picking]{
\includegraphics[width=\env]{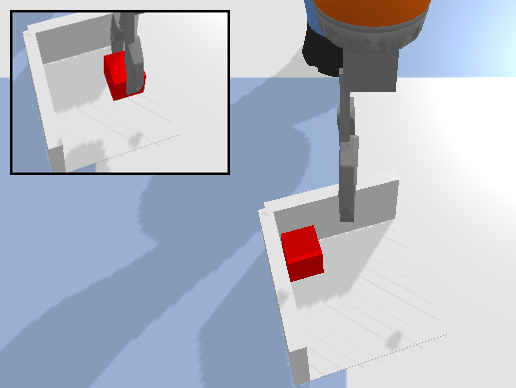}
\label{fig:close_corner}
}\\
\subfloat[Drawer Opening]{
\includegraphics[width=\env]{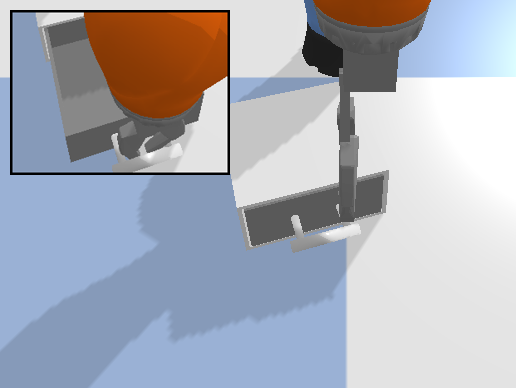}
\label{fig:close_drawer}
}
\subfloat[Object Grasping]{
\includegraphics[width=\env]{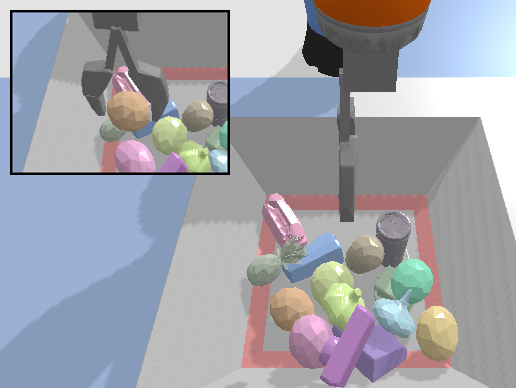}
\label{fig:close_grasp}
}
\caption{The close-loop environments. The window on the top-left corner of each sub-figure shows the goal state of the task.}
\label{fig:close_loop_envs}
\end{figure}

\subsection{Close-loop Environments}
The close-loop environments require the agent to control the delta pose of the 
end-effector, allowing the agent more control and enabling us to solve more 
contact-rich tasks. These environments have a similar setup to the open-loop 
domain but to avoid the occlusion caused by the arm, we instead use two side-view
cameras pointing to the workspace (Figure~\ref{fig:close_ws}). The heightmap 
$I$ is generated by first acquiring a fused point cloud from the two cameras
(Figure~\ref{fig:close_pc}) and then performing an orthographic projection
(Figure~\ref{fig:close_obs}). This orthographic projection is centered with 
respect to the gripper. In practice, this process can be replaced by putting a 
simulated orthographic camera at the position of the gripper to speed up the 
simulation. The state space is defined as a tuple $s=(I, g)\in S$, where
$g\in\{\textsc{holding}, \textsc{empty}\}$ is the gripper state indicating 
if there is an object being held by gripper. The action space is defined
as the cross product of the gripper open width $A_\lambda$ and 
the delta motion of the gripper $A_\delta$, $A=A_\lambda \times A_\delta$. 

Two different action spaces are available for $A_\delta$: 
$A_\delta\in \{A_{\delta}^{xyz}, A_{\delta}^{xyz\theta}\}$. In $A_{\delta}^{xyz}$, 
the robot controls the change of the $x, y, z$ position of the gripper, where 
the top-down orientation $\theta$ is fixed. In $A_{\delta}^{xyz\theta}$, the robot
controls the change of the $x, y, z$ position and the top-down orientation
$\theta$ of the gripper. 
Figure~\ref{fig:close_loop_envs} shows the 10 close-loop environments. We provide a 
default sparse reward function for all environments. 
See Appendix~\ref{appendix:close_envs} for a detailed description of the tasks. 

\section{Benchmark}

BulletArm provides a set of 5 benchmarks covering the various environments and action
spaces (Table \ref{tab:benchmarks}). In this section, we detail the Open-Loop 3D 
Benchmark and the Close-Loop 4D Benchmark. See Appendix \ref{appendix:other_benchmark} 
for the other three benchmarks. 

\begin{table}[t]
\setlength{\tabcolsep}{4pt}
\centering
\begin{tabular}{c c c}
\toprule
Benchmark & Environments & Action Space \\
\midrule
Open-Loop 2D Benchmark & open-loop environments with fixed orientation &  $A_g\times A^{xy}$\\
Open-Loop 3D Benchmark & open-loop environments with random orientation &  $A_g\times A^{xy\theta}$\\
Open-Loop 6D Benchmark & open-loop environments 6DoF extensions &  $A_g\times A^{\SE(3)}$\\
\midrule
Close-Loop 3D Benchmark & close-loop environments with fixed orientation & $A_\lambda \times A_\delta^{xyz}$\\
Close-Loop 4D Benchmark & close-loop environments with random orientation & $A_\lambda \times A_\delta^{xyz\theta}$\\
\bottomrule
\end{tabular}
\caption{The five benchmarks in our work include three open-loop benchmarks and two close-loop benchmarks. `fixed orientation' and `random orientation' indicate whether the objects in the environments will be initialized with a fixed orientation or random orientation.}
\label{tab:benchmarks}
\end{table}

\begin{table}[t]
    \centering
    \setlength{\tabcolsep}{4pt}
    \begin{tabular}{ccccccccccccc}
    \toprule
    Task & \rotatebox{90}{Block Stacking} & \rotatebox{90}{House Building 1} & \rotatebox{90}{House Building 2} & \rotatebox{90}{House Building 3} & \rotatebox{90}{House Building 4} & \rotatebox{90}{\makecell{Improvise House Building 2}} &  \rotatebox{90}{\makecell{Improvise House Building 3}} & \rotatebox{90}{Bin Packing} & \rotatebox{90}{Bottle Arrangement} & \rotatebox{90}{Box Palletizing} & \rotatebox{90}{Covid Test} & \rotatebox{90}{Object Grasping}\\
    \midrule
    Number of Objects & 4 & 4 & 3 & 4 & 6 & 3 & 4 & 8 & 6 & 18 & 6 & 15\\
    \midrule
    Optimal Number of Steps & 6 & 6 & 4 & 6 & 10 & 4 & 6 & 16 & 12 & 36 & 18 & 1\\
    \midrule
    Max Number of Steps & 10 & 10 & 10 & 10 & 20 & 10 & 10 & 20 & 20 & 40 & 30 & 1\\
    \bottomrule
    \end{tabular}
    \caption{The number of objects, optimal number of steps per episode, and max number of steps per episode in our Open-Loop 3D benchmark experiments}
    \label{tab:open_benchmark}
\end{table}

\subsection{Open-Loop 3D Benchmark}
\label{sec:open_ben}

\begin{figure}[t]
\includegraphics[width=\linewidth]{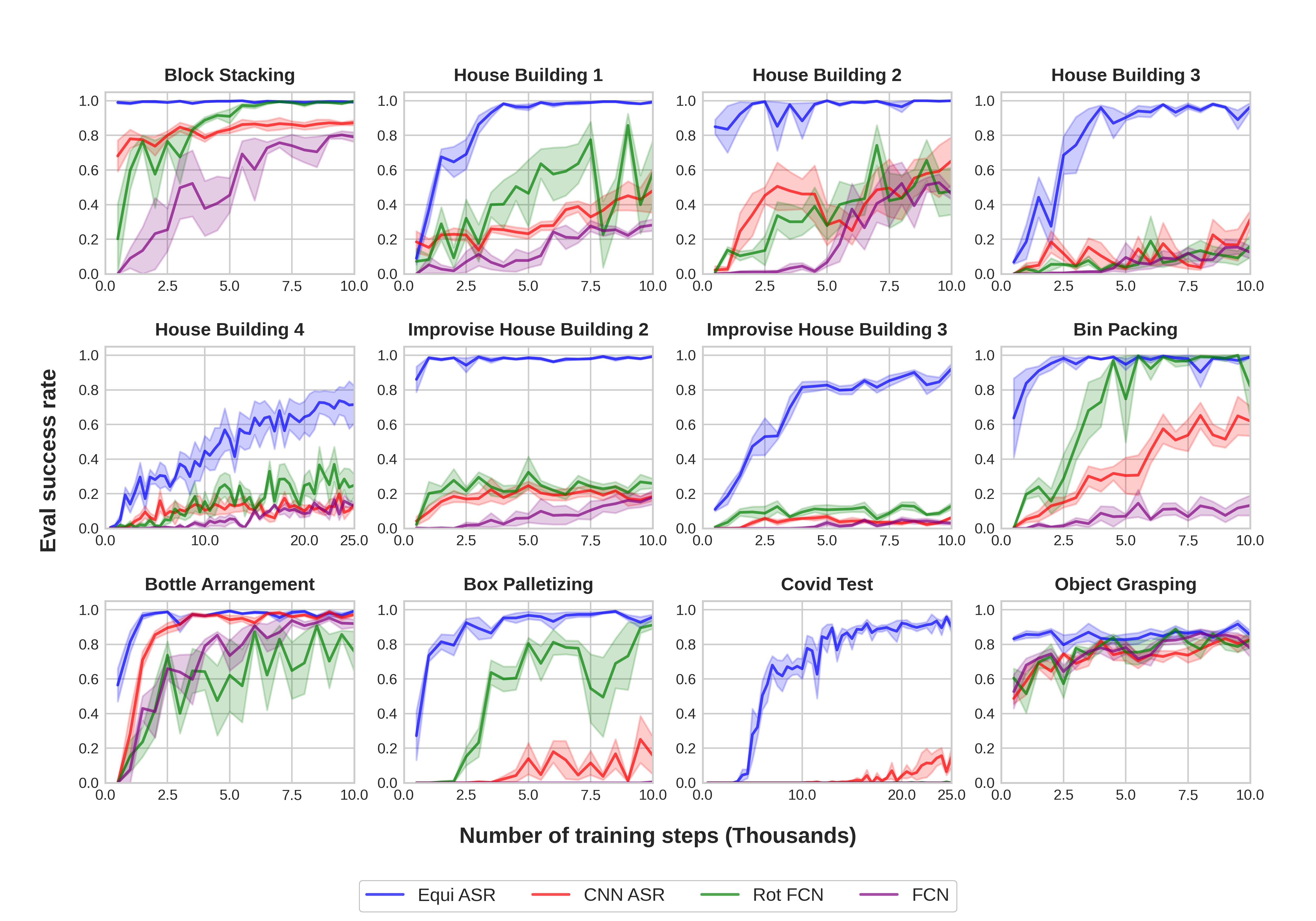}
\centering
\caption{The Open-Loop 3D Benchmark results. The plots show the evaluation performance of the greedy policy in terms of the task success rate. The evaluation is performed every 500 training steps. Results are averaged over four runs. Shading denotes standard error.}
\label{fig:open_exp}
\end{figure}

In the Open-Loop 3D Benchmark, the agent needs to solve the open-loop tasks shown
in Figure~\ref{fig:open_loop_envs} using the $A_g\times A^{xy\theta}$ action space
(see Section~\ref{sec:open_envs}). We provide a set of baseline algorithms that 
explicitly control $(x, y, \theta)\in A^{xy\theta}$ and select the gripper 
motion using the following heuristic: a $\textsc{pick}$ action will be executed 
if $g=\textsc{empty}$ and a $\textsc{place}$ action will be executed if 
$g=\textsc{holding}$. The baselines include: (1) DQN \cite{dqn}, (2) ADET \cite{adet},
(3) DQfD \cite{dqfd}, and (4) SDQfD \cite{asr}. The network architectures for these 
different methods can be used interchangeably. We provide the following network 
architectures:
\begin{enumerate}
    \item CNN ASR~\cite{asr}: A two-hierarchy architecture that selects $(x, y)$
          and $\theta$ sequentially.
    \item Equivariant ASR (Equi ASR)~\cite{equi_asr}: Similar to ASR, but instead 
          of using conventional CNNs, equivariant steerable CNNs \cite{steerable_cnns,e2cnn}
          are used to capture the rotation symmetry of the tasks.
    \item FCN: a Fully Convolutional Network (FCN)~\cite{fcn} which outputs a $n$ 
          channel action-value map for each discrete rotation.
    \item Equivariant FCN~\cite{equi_asr}: Similar to FCN, but instead of using 
           conventional CNNs, equivariant steerable CNNs are used. 
    \item Rot FCN \cite{zeng_pick,zeng_push}: A FCN with 1-channel input and 
          output, the rotation is encoded by rotating the input and output for each
          $\theta$.
\end{enumerate}

In this section, we show the performance of SDQfD (which is shown to be better than
DQN, ADET, and DQfD~\cite{asr}. See the performance of DQN, ADET and DQfD in Appendix~\ref{appendix:open_3d_extra}) equipped with CNN ASR, Equi ASR, FCN, and Rot FCN.
We evaluate SDQfD in the 12 environments shown in Figure~\ref{fig:open_loop_envs}. 
Table~\ref{tab:open_benchmark} shows the number of objects, the optimal number of steps
per episode, and the max number of steps per episode in the open-loop benchmark 
experiments. Before the start of training, 200 (500 for Covid Test) episodes of expert
data are populated in the replay buffer. Figure~\ref{fig:open_exp} shows the results.
Equivariant ASR (blue) has the best performance across all environments, then 
Rot FCN (green) and CNN ASR (red), and finally FCN (purple). Notice that Equivariant 
ASR is the only method that is capable of solving the most challenging tasks
(e.g., Improvise House Building 3 and Covid Test).

\subsection{Close-Loop 4D Benchmark}
\label{sec:close_ben}

\begin{table}[t]
    \centering
    \setlength{\tabcolsep}{4pt}
    \begin{tabular}{ccccccccccc}
    \toprule
    Task & \rotatebox{90}{Block Reaching} & \rotatebox{90}{Block Picking} & \rotatebox{90}{Block Pushing} & \rotatebox{90}{Block Pulling} & \rotatebox{90}{Block in Bowl} & \rotatebox{90}{Block Stacking} &  \rotatebox{90}{House Building} & \rotatebox{90}{Corner Picking} & \rotatebox{90}{Drawer Opening} & \rotatebox{90}{Object Grasping}\\
    \midrule
    Number of Objects & 1 & 1 & 1 & 2 & 2 & 2 & 2 & 1 & 1 & 5\\
    \midrule
    Optimal Number of Steps & 3 & 7 & 7 & 7 & 11 & 12 & 12 & 14 & 9 & 7\\
    \midrule
    Max Number of Steps & 50 & 50 & 50 & 50 & 50 & 50 & 50 & 50 & 50 & 50\\
    \bottomrule
    \end{tabular}
    \caption{The number of objects, optimal number of steps per episode, and max number of steps per episode in our Close-Loop 4D Benchmark experiments.}
    \label{tab:close_benchmark}
\end{table}

\begin{figure}[t]
\includegraphics[width=\linewidth]{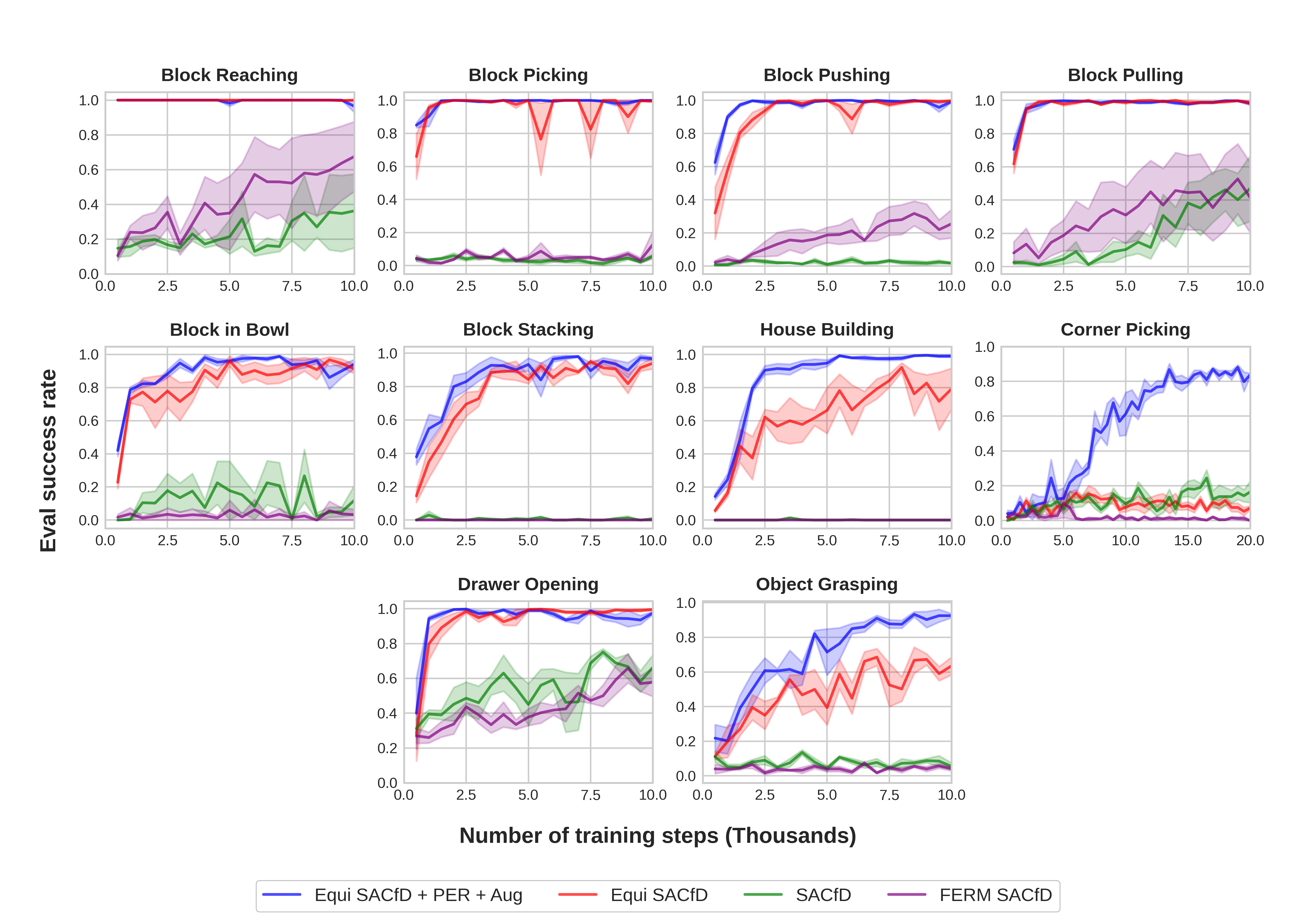}
\centering
\caption{The Close-Loop 4D benchmark results. The plots show the evaluation performance of the greedy policy in terms of the task success rate. The evaluation is performed every 500 training steps. Results are averaged over four runs. Shading denotes standard error.}
\label{fig:close_exp}
\end{figure}

The Close-Loop 4D Benchmark requires the agent to solve the close-loop tasks shown
in Figure~\ref{fig:close_loop_envs} in the 5-dimensional action space of 
$(\lambda, x, y, z, \theta) \in A_\lambda \times A_{\delta}^{ xyz\theta} \subset \mathbb{R}^5$,
where the agent controls the positional displacement of the gripper 
($x, y, z$), the rotational displacement of the gripper along the $z$ axis ($\theta$),
and the open width of the gripper ($\lambda$). We provide the following baseline 
algorithms: (1) SAC~\cite{sac}, (2) Equivariant SAC~\cite{equi_rl}, (3) RAD~\cite{rad} SAC: 
SAC with data augmentation, (4) DrQ~\cite{drq} SAC: Similar to (3), but 
performs data augmentation when calculating the $Q$-target and the loss, and (5) 
FERM~\cite{ferm}: A Combination of SAC and contrastive learning~\cite{curl} using data
augmentation. Additionally, we also provide a variation of SAC, SACfD~\cite{equi_rl}, 
that applies an auxiliary L2 loss towards the expert action to the actor network. 
SACfD can also be used in combination with (2), (3), and (4) to form Equivariant SACfD,
RAD SACfD, DrQ SACfD, and FERM SACfD.

In this section, we show the performance of SACfD, Equivariant SACfD (Equi SACfD), 
Equivariant SACfD using Prioritized Experience Replay (PER~\cite{per}) and 
data augmentation (Equi SACfD + PER + Aug), and FERM SACfD. 
(See Appendix~\ref{appendix:close_4d_extra} for the performance of RAD SACfD and 
DrQ SACfD.) 
We use a continuous action space where 
$x, y, z\in [-0.05m, 0.05m], \theta \in [-\frac{\pi}{4}, \frac{\pi}{4}], \lambda\in [0, 1]$.
We evaluate the various methods in the 10 environments shown in
Figure~\ref{fig:close_loop_envs}. Table~\ref{tab:close_benchmark} shows the number 
of objects, the optimal number of steps for solving each task, and the maximal
number of steps for each episode. In all tasks, we pre-load 100 episodes of expert
demonstrations in the replay buffer.

Figure~\ref{fig:close_exp} shows the performance of the baselines. Equivariant SACfD 
with PER and data augmentation (blue) has the best overall performance followed by 
standard Equivariant SACfD (red). The equivariant algorithms show a significant 
improvement when compared to the other algorithms which do not encode rotation symmetry, 
i.e. CNN SACfD and FERM SACfD.

\section{Conclusions}
In this paper, we present BulletArm, a novel benchmark and learning environment aimed
at robotic manipulation. By providing a number of manipulation tasks alongside 
our baseline algorithms, we hope to encourage more direct comparisons between new
methods. This type of standardization through direct comparison has been a key aspect 
of improving research in deep learning methods for both computer vision and 
reinforcement learning. We aim to maintain and improve this framework for the 
foreseeable future adding new features, tasks, and baseline algorithms over time. 
An area of particular interest for us is to extend the existing suite of tasks to include
more dynamic environments where the robot is tasked with utilizing tools to accomplish various tasks.
We hope that with the aid of the community, this repository will grow over time to 
contain both a large number of interesting tasks and state-of-the-art baseline algorithms. 

\section*{Acknowledgments}
This work is supported in part by NSF 1724257, NSF 1724191, NSF 1763878, NSF 1750649, and NASA 80NSSC19K1474.

%
%
\bibliographystyle{abbrv}
\bibliography{main}

\clearpage
\appendix

\section{Detail Description of Environments}
\subsection{Open-Loop Environments}
\label{appendix:open_envs}
\subsubsection{Block Stacking}
In the Block Stacking environment (Figure~\ref{fig:open_stack}), there are $N$ cubic blocks with a size of $3cm\times 3cm\times 3cm$. The blocks are randomly initialized in the workspace. The goal of this task is to stack all blocks in a stack. An optimal policy requires $2(N-1)$ steps to finish this task. The number of blocks $N$ is configurable. By default, $N=4$, and the maximal number of steps per episode is 10.

\subsubsection{House Building 1} In the House Building 1 environment (Figure~\ref{fig:open_h1}), there are $N-1$ cubic blocks with a size of $3cm\times 3cm\times 3cm$ and one triangle block with a bounding box size of around $3cm\times 3cm\times 3cm$. The blocks are randomly initialized in the workspace. The goal of this task is to first form a stack using the $N-1$ cubic blocks, then place the triangle block on top of the stack. An optimal policy requires $2(N-1)$ steps to finish this task. The number of blocks $N$ is configurable. By default, $N=4$, and the maximal number of steps per episode is 10.

\subsubsection{House Building 2} In the House Building 2 environment (Figure~\ref{fig:open_h2}), there are two cubic blocks with a size of $3cm\times 3cm\times 3cm$, and a roof block with a bounding box size of around $12cm\times 3cm\times 3cm$. The blocks are randomly initialized in the workspace. The goal of this task is to place the two cubic blocks next to each other, then place the roof block on top of the two cubic blocks. An optimal policy requires 4 steps to finish this task. The default maximal number of steps per episode is 10.

\subsubsection{House Building 3} In the House Building 3 environment (Figure~\ref{fig:open_h3}), there are two cubic blocks with a size of $3cm\times 3cm\times 3cm$, one cuboid block with a size of $12cm\times 3cm\times 3cm$, and a roof block with a bounding box size of around $12cm\times 3cm\times 3cm$. The blocks are randomly initialized in the workspace. The goal of this task is to first place the two cubic blocks next to each other, place the cuboid block on top of the two cubic blocks, then place the roof block on top of the cuboid block. An optimal policy requires 6 steps to finish this task. The default maximal number of steps per episode is 10.

\subsubsection{House Building 4} In the House Building 4 environment (Figure~\ref{fig:open_h4}), there are four cubic blocks with a size of $3cm\times 3cm\times 3cm$, one cuboid block with a size of $12cm\times 3cm\times 3cm$, and a roof block with a bounding box size of around $12cm\times 3cm\times 3cm$. The blocks are randomly initialized in the workspace. The goal of this task is to first place two cubic blocks next to each other, place the cuboid block on top of the two cubic blocks, place another two cubic blocks on top of the cuboid block, then place the roof block on top of the two cubic blocks. An optimal policy requires 10 steps to finish this task. The default maximal number of steps per episode is 20.

\begin{wrapfigure}[7]{r}{0.5\textwidth}
\centering
\includegraphics[width=\linewidth]{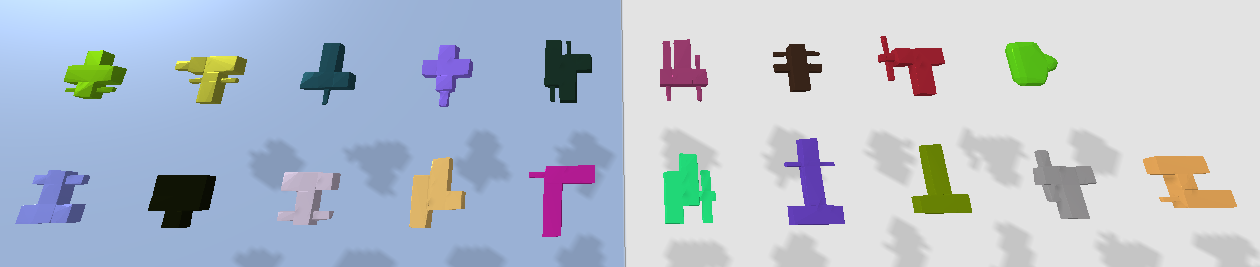}
\caption{The object set in the Improvise House Building 2 and Improvise House Building 3 environment.}
\label{fig:improvise_object_set}
\end{wrapfigure}
\subsubsection{Improvise House Building 2} In the Improvise House Building 2 environment (Figure~\ref{fig:open_imh2}), there are two random blocks and a roof block. The shapes of the random blocks are sampled from Figure~\ref{fig:improvise_object_set}. The blocks are randomly initialized in the workspace. The goal of this task is to place the two random blocks next to each other, then place the roof block on top of the two random blocks. An optimal policy requires 4 steps to finish this task. The default maximal number of steps per episode is 10.

\subsubsection{Improvise House Building 3} In the Improvise House Building 3 environment (Figure~\ref{fig:open_imh3}), there are two random blocks, a cuboid block, and a roof block. The shapes of the random blocks are sampled from Figure~\ref{fig:improvise_object_set}. The blocks are randomly initialized in the workspace. The goal of this task is to place the two random blocks next to each other, place the cuboid block on top of the two random blocks, then place the roof block on top of the cuboid block. An optimal policy requires 6 steps to finish this task. The default maximal number of steps per episode is 10.

\begin{wrapfigure}[8]{r}{0.4\textwidth}
\vspace{-0.8cm}
\centering
\includegraphics[width=\linewidth]{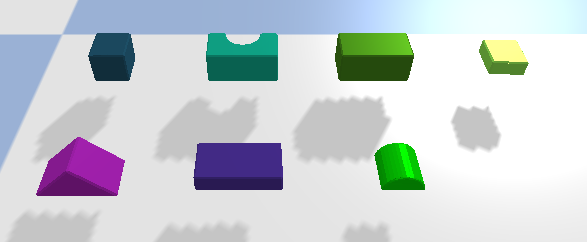}
\caption{The object set in the Bin Packing environment.}
\label{fig:pack_object_set}
\end{wrapfigure}
\subsubsection{Bin Packing} In the Bin Packing task (Figure~\ref{fig:open_pack}), $N$ objects and a bin are randomly placed in the workspace. The shapes of the objects are randomly sampled from Figure~\ref{fig:pack_object_set} (Object models are derived from~\cite{zeng_push}) with a maximum size of $8cm\times4cm\times4cm$ and a minimum size of $4cm\times 4cm\times 2cm$.  The bin has a size of $17.6cm\times 14.4cm\times 8cm$. The goal of this task is to pack all $N$ objects in the bin. An optimal policy requires $2N$ steps to finish the task. The number of objects $N$ is configurable. By default, $N=8$, and the maximal number of steps per episode is 20. 

\begin{wrapfigure}[8]{r}{0.4\textwidth}
\vspace{-0.8cm}
\centering
\includegraphics[width=\linewidth]{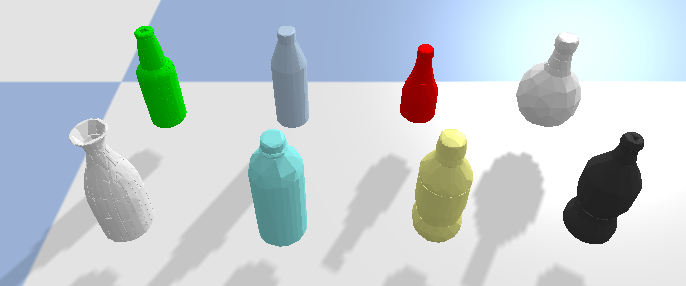}
\caption{The object set in the Bottle arrangement environment.}
\label{fig:bottle_object_set}
\end{wrapfigure}
\subsubsection{Bottle Arrangement} In the Bottle Arrangement task (Figure~\ref{fig:open_bottle}), six bottles with random shapes (sampled from 8 different shapes shown in Figure~\ref{fig:bottle_object_set}. The bottle shapes are generated from the 3DNet dataset~\cite{3dnet}. The sizes of each bottle are around $5cm\times 5cm\times 14cm$), and a tray with a size of $24cm\times 16cm\times 5cm$ are randomly placed in the workspace. The goal is to arrange all six bottles in the tray. An optimal policy requires 12 steps to finish this task. By default, the maximal number of steps per episode is 20.

\subsubsection{Box Palletizing} In the Box Palletizing task (Figure~\ref{fig:open_box}) (some object models are derived from~\cite{transporter}), a pallet with a size of $23.2cm\times 19.2cm\times 3cm$ is randomly placed in the workspace. The goal is to stack $N$ boxes with a size of $7.2cm\times 4.5cm\times 4.5cm$ as shown in Fig~\ref{fig:open_box}. At the beginning of each episode and after the agent correctly places a box on the pallet, a new box will be randomly placed in the empty workspace. An optimal policy requires $2N$ steps to finish this task. The number of boxes $N$ is configurable (6, 12, or 18). By default, $N=18$, and the maximal number of steps per episode is 40.

\begin{figure}[t]
\centering
\subfloat[]{
\includegraphics[width=0.15\linewidth]{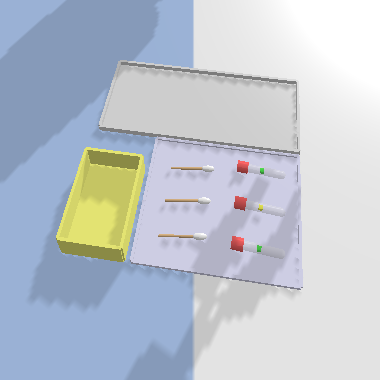}
}
\subfloat[]{
\includegraphics[width=0.15\linewidth]{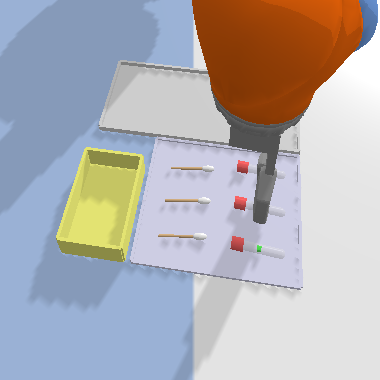}
}
\subfloat[]{
\includegraphics[width=0.15\linewidth]{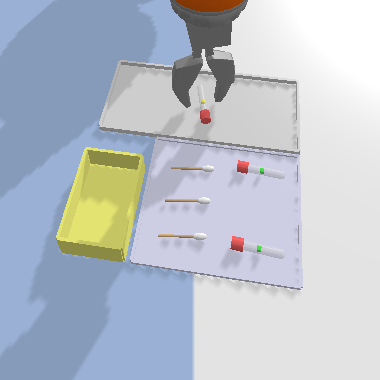}
}
\subfloat[]{
\includegraphics[width=0.15\linewidth]{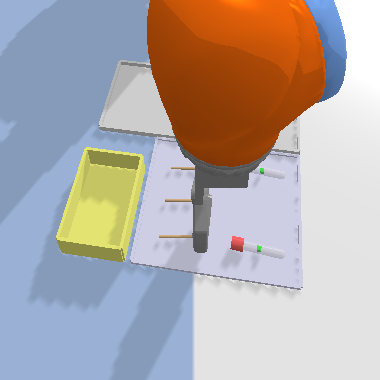}
}\\
\subfloat[]{
\includegraphics[width=0.15\linewidth]{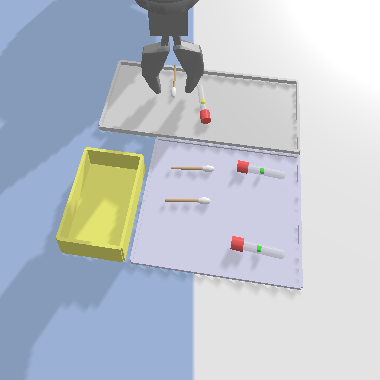}
}
\subfloat[]{
\includegraphics[width=0.15\linewidth]{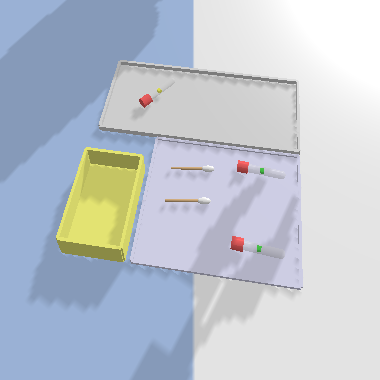}
}
\subfloat[]{
\includegraphics[width=0.15\linewidth]{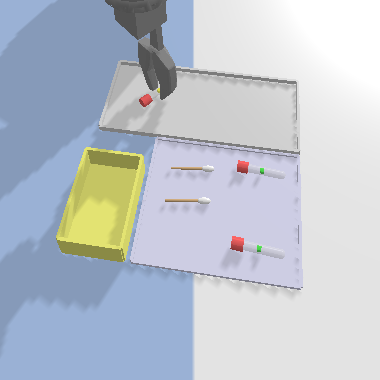}
}
\subfloat[]{
\includegraphics[width=0.15\linewidth]{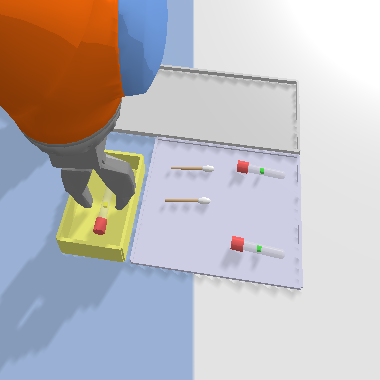}
}
\caption{An example of one COVID test process.}
\label{fig:covid}
\end{figure}

\subsubsection{Covid Test} In the Covid Test task (Figure~\ref{fig:open_covid}), there is a new tube box (purple), a test area (gray), and a used tube box (yellow) placed arbitrarily in the workspace but adjacent to one another. Three swabs with a size of $7cm \times 1cm \times 1cm$ and three tubes with a size of $8cm \times 1.7cm \times 1.7cm$ are initialized in the new tube box. To supervise a COVID test, the robot needs to present a pair of a new swab and a new tube from the new tube box to the test area. The simulator simulates the user testing COVID by putting the swab into the tube and randomly placing the used tube in the test area. Then the robot needs to re-collect the used tube into the used tube box. See one example of this process in Figure~\ref{fig:covid}. Each episode includes three rounds of COVID test. An optimal policy requires 18 steps to finish this task. By default, the maximal number of steps per episode is 30.

\begin{figure}[t]
\centering
\includegraphics[width=0.5\linewidth]{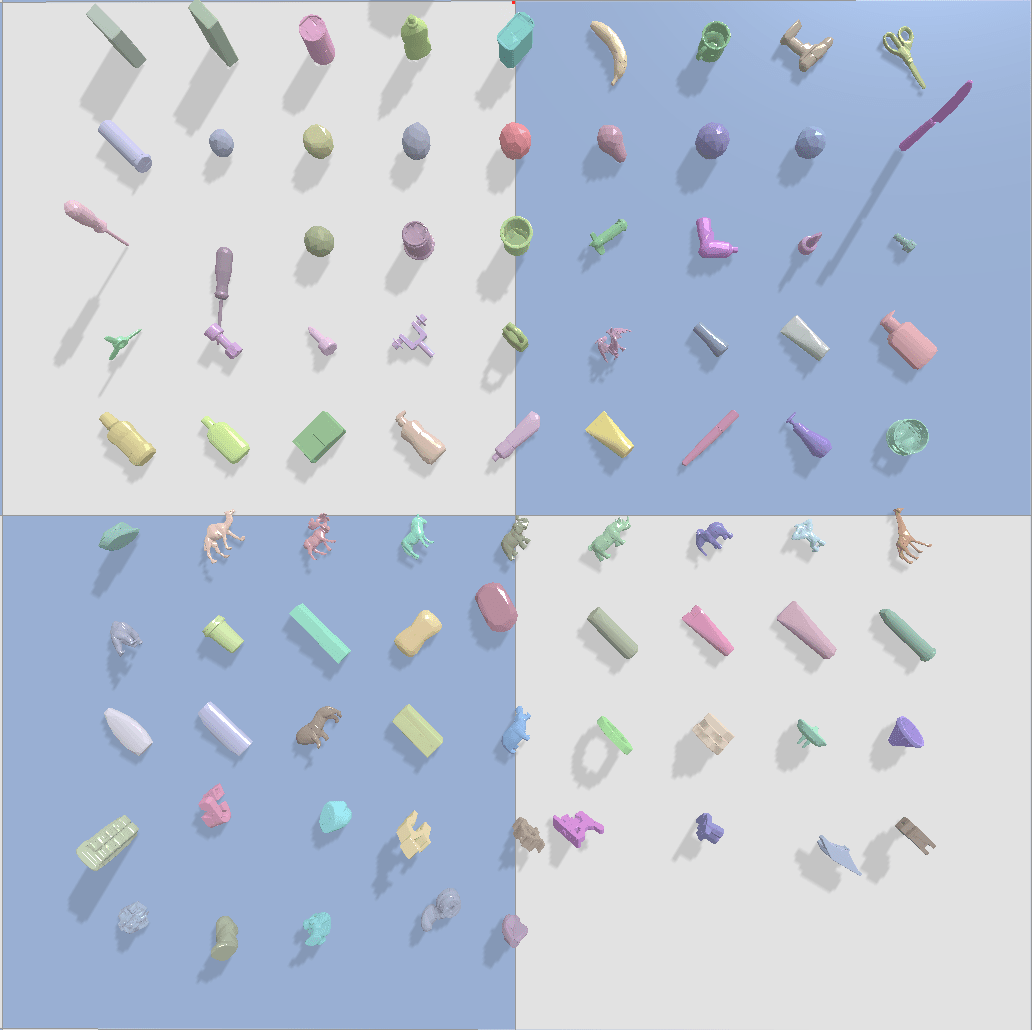}
\caption{The object set in the Object Grasping environment.}
\label{fig:grasp_object_set}
\end{figure}

\subsubsection{Object Grasping}
In the Object Grasping task (Figure~\ref{fig:open_grasp}), the robot needs to grasp an object from a clutter of at most $N$ objects. At the start of training, $N$ random objects are initialized with random position and orientation. The shapes of the objects are randomly sampled from the object set shown in Figure~\ref{fig:grasp_object_set}. The object set contains 86 objects derived from the GraspNet1B~\cite{graspnet_1b} dataset. Every time the agent successfully grasps all $N$ objects, the environment will re-generate $N$ random objects with random positions and orientations. The maximal number of steps per episode is 1. The number of objects $N$ in this environment is configurable. By default, there will be 15 objects.

\subsection{Close-Loop Environments}
\label{appendix:close_envs}
\subsubsection{Block Reaching} In the Block Reaching environment (Figure~\ref{fig:close_reach}), there is a cubic block with a size of $5cm\times 5cm\times 5cm$. The block is randomly initialized in the workspace. The goal of this task is to move the gripper towards the block such that the distance of the fingertip and the block is within $3cm$. By default, the maximal number of steps per episode is 50.

\subsubsection{Block Picking} In the Block Picking environment (Figure~\ref{fig:close_pick}), there is a cubic block with a size of $5cm\times 5cm\times 5cm$. The block is randomly initialized in the workspace. The goal of this task is to grasp the block and raise the gripper such that the gripper is $15cm$ above the ground. By default, the maximal number of steps per episode is 50.

\subsubsection{Block Pushing} In the Block Pushing environment (Figure~\ref{fig:close_push}), there is a cubic block with a size of $5cm\times 5cm\times 5cm$ and a goal area with a size of $9cm\times 9cm$. The block and the goal area are randomly initialized in the workspace. The goal of this task is to push the block such that the distance between the block's center and the goal's center is within $5cm$. By default, the maximal number of steps per episode is 50.

\subsubsection{Block Pulling} In the Block Pulling environment (Figure~\ref{fig:close_pull}), there are two cuboid blocks with a size of $8cm\times 8cm\times 5cm$. The blocks are randomly initialized in the workspace. The goal of this task is to pull one of the two blocks such that it makes contact with another block. By default, the maximal number of steps per episode is 50.

\subsubsection{Block in Bowl} In the Block in Bowl environment (Figure~\ref{fig:close_bowl}), there is a cubic block with a size of $5cm\times 5cm\times 5cm$, and a Bowl with a bounding box size of $16cm\times 16cm \times 7cm$. The block and the bowl are randomly initialized in the workspace. The goal of this task is to pick up the block and place it inside the bowl. By default, the maximal number of steps per episode is 50.

\subsubsection{Block Stacking} In the Block Stacking environment (Figure~\ref{fig:close_stack}), there are N cubic blocks with a size of $5cm\times 5cm\times 5cm$. The blocks are randomly initialized in the workspace. The goal of this task is to form a stack using the $N$ blocks. By default, $N=2$, the maximum number of steps per episode is 50.

\subsubsection{House Building} In the House Building environment (Figure~\ref{fig:close_h1}), there are $N-1$ cubic blocks with a size of $5cm\times 5cm\times 5cm$ and one triangle with a bounding box size of $5cm\times 5cm\times 5cm$. The blocks are randomly initialized in the workspace. The goal of this task is to first form a stack using the $N-1$ cubic blocks, then place the triangle block on top. By default, $N=2$, the maximum number of steps per episode is 50.

\subsubsection{Corner Picking} In the Corner Picking environment (Figure~\ref{fig:close_corner}), there is a cubic block with a size of $5cm\times 5cm\times 5cm$ and a corner formed by two walls. The poses of the block and the corner are randomly initialized with a fixed relative pose between them so that the block is right next to the two walls. The wall is fixed in the workspace and not movable. The goal of this task is to nudge the block out from the corner and then pick it up at least $15cm$ above the ground. By default, the maximum number of steps per episode is 50.

\subsubsection{Drawer Opening} In the Drawer Opening environment (Figure~\ref{fig:close_drawer}), there is a drawer with a random pose in the workspace. The outer part of the drawer is fixed and not movable. The goal of this task is to pull the drawer handle to open the drawer. By default, the maximum number of steps per episode is 50.

\subsubsection{Object Grasping} In the Object Grasping task (Figure~\ref{fig:open_grasp}), the robot needs to grasp an object from a clutter of at most $N$ objects. At the start of training, $N$ random objects are initialized with random position and orientation. The shapes of the objects are randomly sampled from the object set shown in Figure~\ref{fig:grasp_object_set}. The object set contains 86 objects derived from the GraspNet1B~\cite{graspnet_1b} dataset. Every time the agent successfully grasps all $N$ objects, the environment will re-generate $N$ random objects with random positions and orientations. If an episode terminates with any remaining objects in the bin, the objects will not be re-initialized. The goal of this task is to grasp any object and lift it such that the gripper is at least $0.15m$ above the ground. The number of objects $N$ in this environment is configurable. By default, there will be 5 objects, and the maximum number of steps per episode is 50.

\section{List of Configurable Environment Parameters}
\label{appendix:parameters}

\begin{table}[t]
\newcolumntype{s}{>{\hsize=.4\hsize}X}
\centering
\begin{tabularx}{\textwidth}{ssX}
\toprule
Parameter & Example & Description \\
\midrule
robot & kuka & the robot to use in the experiment. \\
\midrule
action\_sequence & pxyzr & The action space. `pxyzr' means the action space a 5-vector, including the gripper action (p), the position of the gripper (x, y, z), and its top-down rotation (r).\\
\midrule
workspace & array([[0.25, 0.65], [-0.2, 0.2], [0, 1]]) & The workspace in terms of the range in x, y, and z.\\
\midrule
object\_scale\_range & 0.6 & The scale of the size of the objects in the environment. \\
\midrule
max\_steps & 10 & The maximal steps per episode.\\
\midrule
num\_objects & 1 & The number of objects in the environment.\\
\midrule
obs\_size & 128 & The pixel size of the observation $I$.\\
\midrule
in\_hand\_size & 24 & The pixel size of the in-hand image $H$.\\
\midrule
fast\_mode & True & If True, teleport the arm when possible to speed up the simulation.\\
\midrule
render & False & If True, render the PyBullet GUI.\\
\midrule
random\_orientation & True & If True, the objects in the environments will be initialized with random orientations. \\
\midrule
half\_rotation & True & If True, constrain the gripper rotation between 0 and $\pi$. \\
\midrule
workspace\_check & point/bounding\_box & Check object out of bound using the object center of mass or the bounding box \\
\midrule
close\_loop\_tray & False & If True, generate a tray like in the Object Grasping (Figure~\ref{fig:close_grasp}) in the close-loop environment. \\
\bottomrule
\end{tabularx}
\caption{List of example configurable parameters in our framework.}
\label{tab:parameters}
\end{table}

Table~\ref{tab:parameters} shows a list of configuration parameters in our framework. 

\section{Open-Loop 6DoF Environments}
\label{appendix:6dof}

\begin{figure}[t]
\captionsetup[subfloat]{format=hang,justification=centering}
\setlength{\env}{0.3\linewidth}
\centering
\subfloat[Ramp Block Stacking]{
\includegraphics[width=\env]{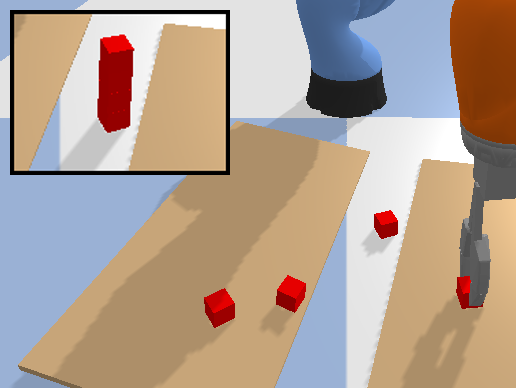}
\label{fig:open_ramp_stack}
}
\subfloat[Ramp House Building 1]{
\includegraphics[width=\env]{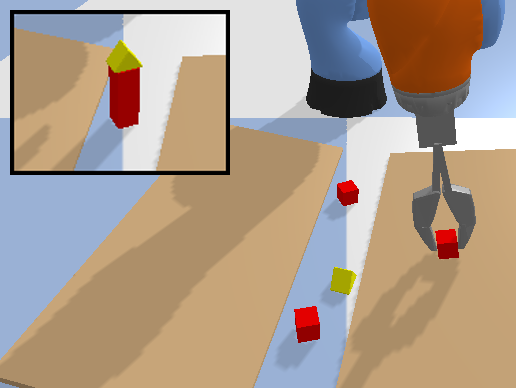}
\label{fig:open_ramp_h1}
}
\subfloat[Ramp House Building 2]{
\includegraphics[width=\env]{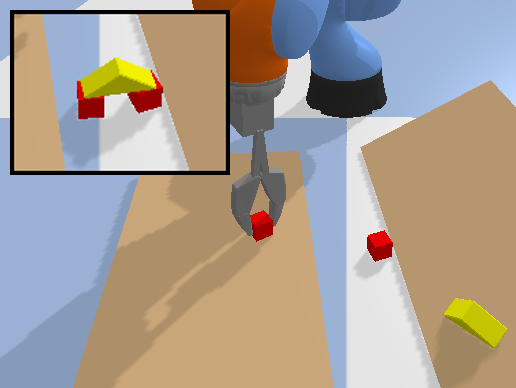}
\label{fig:open_ramp_h2}
}\\
\subfloat[Ramp House Building 3]{
\includegraphics[width=\env]{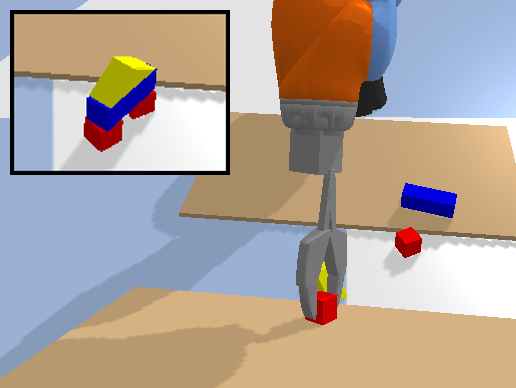}
\label{fig:open_ramp_h3}
}
\subfloat[Ramp House Building 4]{
\includegraphics[width=\env]{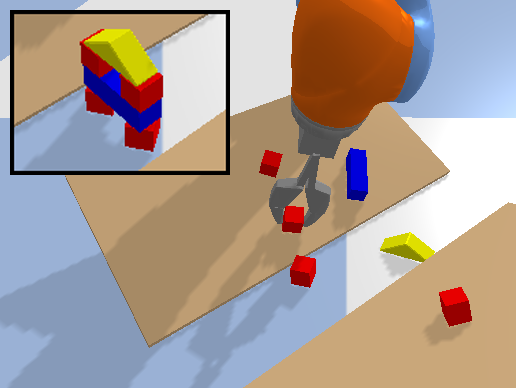}
\label{fig:open_ramp_h4}
}
\subfloat[Ramp Improvise House\\ Building 2]{
\includegraphics[width=\env]{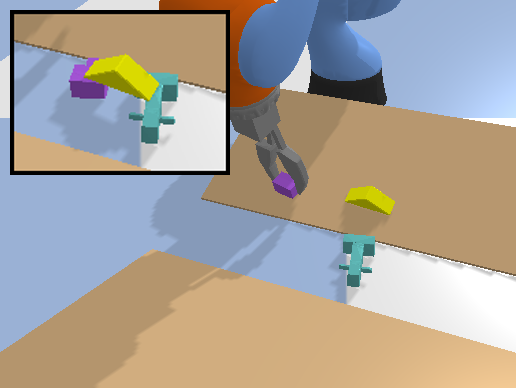}
\label{fig:open_ramp_imh2}
}\\
\subfloat[Ramp Improvise House\\ Building 3]{
\includegraphics[width=\env]{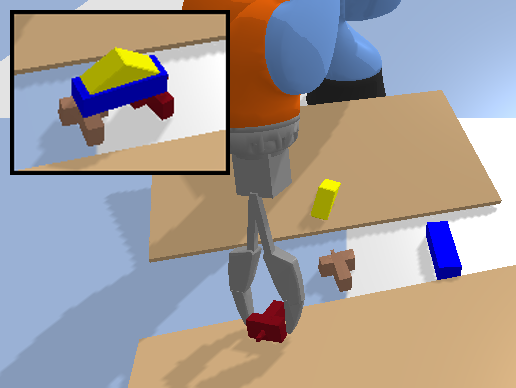}
\label{fig:open_ramp_imh3}
}
\subfloat[Bump House Building 4]{
\includegraphics[width=\env]{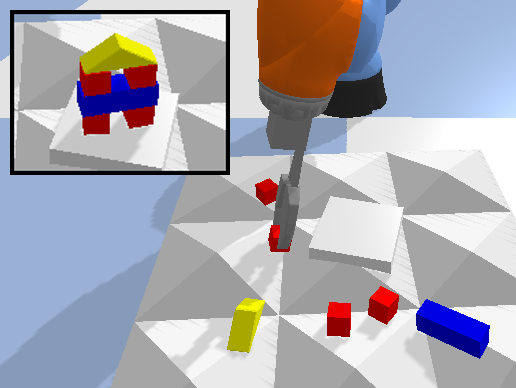}
\label{fig:open_bump_h4}
}
\subfloat[Bump Box Palletizing]{
\includegraphics[width=\env]{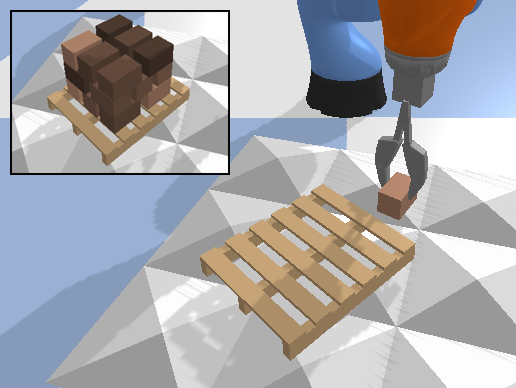}
\label{fig:open_bump_box}
}
\caption{The open-loop 6DoF environments. The window on the top-left corner of each sub-figure shows the goal state of each task.}
\label{fig:open_loop_6d_envs}
\end{figure}

Most of the 6DoF environments mirror those in Figure~\ref{fig:open_loop_envs}, but the workspace is initialized with two ramps in the ramp environments or with a bumpy surface in the bump environments. 

In the ramp environments (Figure~\ref{fig:open_ramp_stack}-Figure~\ref{fig:open_ramp_imh3}), the two ramps are always parallel to each other. The distance between the ramps is randomly sampled between $4cm$ and $20cm$. The orientation of the two ramps along the $z$-axis is randomly sampled between 0 and $2\pi$. The slope of each ramp is randomly sampled between 0 and $\frac{\pi}{6}$. The height of each ramp above the ground is randomly sampled between $0cm$ and $1cm$. In addition, the relevant objects are initialized with random positions and orientations either on the ramps or on the ground.

In the bump environments (Figure~\ref{fig:open_bump_h4} and Figure~\ref{fig:open_bump_box}),  bumpy surface is generated by nine pyramid shapes with a random slop sampled from 0 to $\frac{\pi}{12}$ degrees. The orientation of the bumpy surface along the z-axis is randomly sampled at the beginning of each episode.

\begin{figure}[t]
\includegraphics[width=\linewidth]{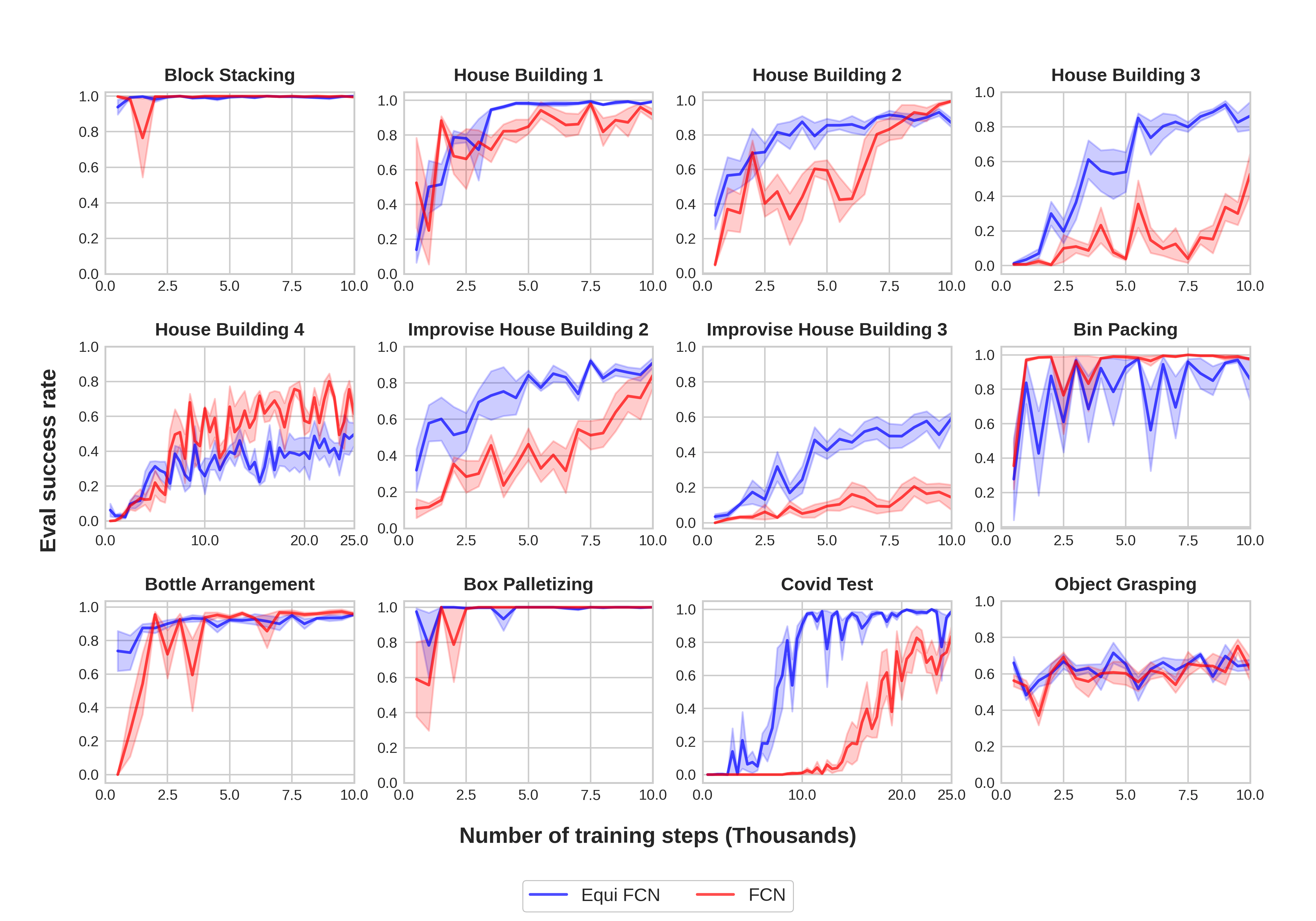}
\centering
\caption{The Open-Loop 2D benchmark result. The plots show the evaluation performance of the greedy policy in terms of the task success rate. The evaluation is performed every 500 training steps. Results are averaged over four runs. Shading denotes standard error.}
\label{fig:open_2d_exp}
\end{figure}

\section{Additional Benchmarks}
\label{appendix:other_benchmark}
This section demonstrates the Open-Loop 2D Benchmark, the Open-Loop 6D Benchmark, and the Close-loop 3D Benchmark (Table~\ref{tab:benchmarks}).
\subsection{Open-Loop 2D Benchmark}
\label{appendix:open_2d_ben}
The Open-Loop 2D Benchmark requires the agent to solve the open-loop tasks in Figure~\ref{fig:open_loop_envs} without random orientations, i.e., all of the objects in the environment will be initialized with a fixed orientation. The action space in this benchmark is $A_g\times A^{xy}$, i.e., the agent only controls the target $(x, y)$ position of the gripper, while $\theta$ is fixed at 0 degree. Other environment parameters mirror the Open-Loop 3D Benchmark in Section~\ref{sec:open_ben}.


Similar as in Section~\ref{sec:open_ben}, we provide DQN, ADET, DQfD, and SDQfD algorithms with FCN and Equivariant FCN (Equi FCN) network architectures (the other architectures do not apply to this benchmark because the agent does not control $\theta$). In this section, we show the performance of SDQfD equipped with FCN and Equivariant FCN. Figure~\ref{fig:open_2d_exp} shows the result. Equivariant FCN (blue) generally shows a better performance compared with standard FCN (red).

\begin{figure}[t]
\includegraphics[width=0.7\linewidth]{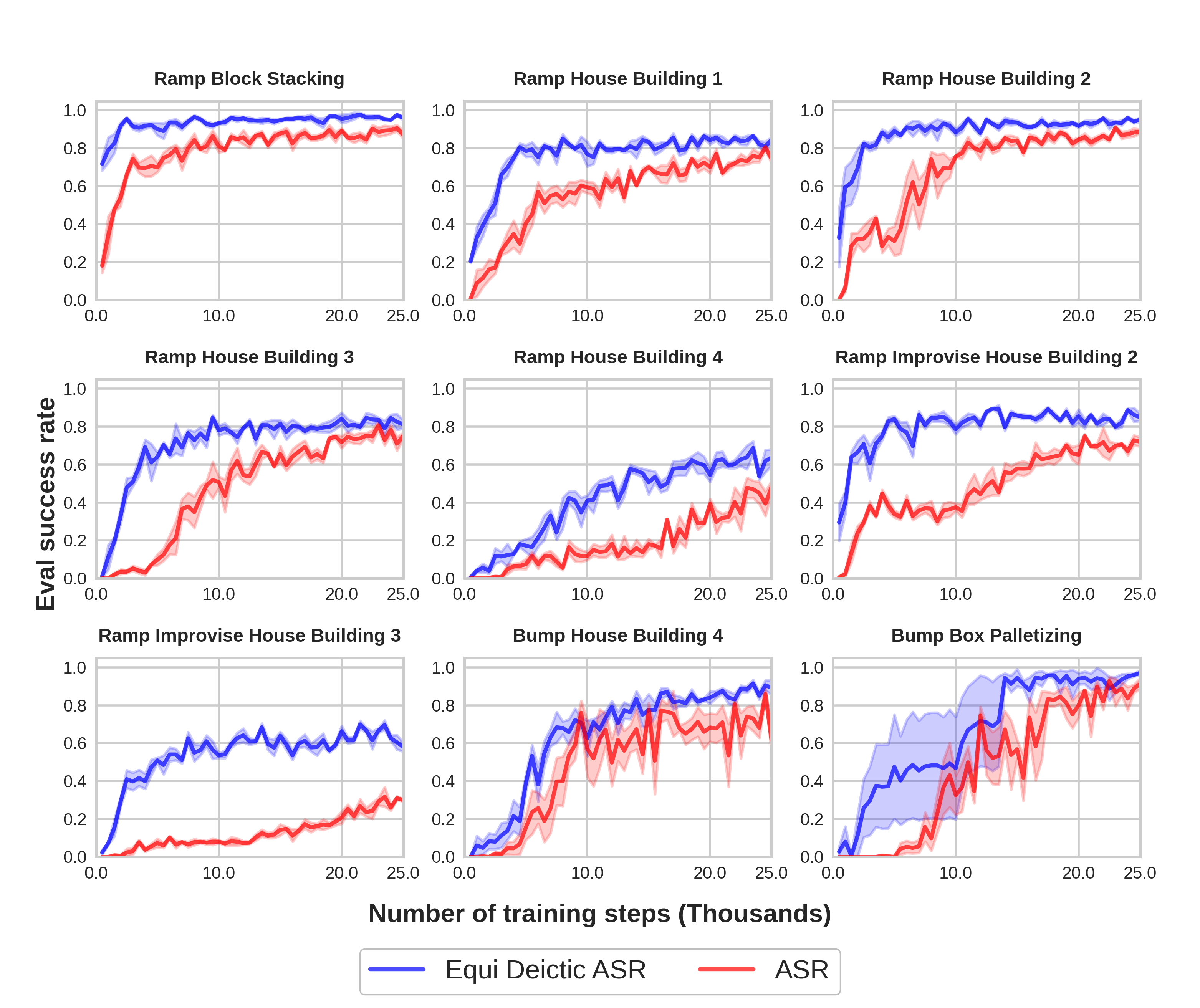}
\centering
\caption{The Open-Loop 6D benchmark result. The plots show the evaluation performance of the greedy policy in terms of the task success rate. The evaluation is performed every 500 training steps. Results are averaged over four runs. Shading denotes standard error.}
\label{fig:open_6d_exp}
\end{figure}
\subsection{Open-Loop 6D Benchmark}
In the Open-Loop 6D Benchmark, the agent needs to solve the open-loop 6DoF environments (Appendix~\ref{appendix:6dof}) in an action space of $A_g\times A^{\SE(3)}$, i.e., the position $(x, y, z)$ of the gripper and the rotation $(\theta, \phi, \psi)$ of the gripper along the $z, y, x$ axes. 

We provide two baselines in this benchmark: 1) ASR~\cite{asr}: a hierarchical approach that selects the actions in a sequence of $((x, y), \theta, z, \phi, \psi)$ using 5 networks; 2) Equivariant Deictic ASR~\cite{equi_asr} (Equi Deictic ASR): similar as 1), but replace the standard networks with equivariant networks and the deictic encoding to improve the sample efficiency. We use 1000 planner episodes for the ramp environments and 200 planner episodes for the bump environments. The in-hand image $H$ in this experiment is a 3-channel orthographic projection image of a voxel grid generated from the point cloud at the previous pick pose. Other environment parameters mirror the Open-Loop 3D Benchmark in Section~\ref{sec:open_ben}.

Figure~\ref{fig:open_6d_exp} shows the results. Equivariant Deictic ASR (blue) demonstrates a stronger performance compared with standard ASR (red).

\begin{figure}[t]
\includegraphics[width=\linewidth]{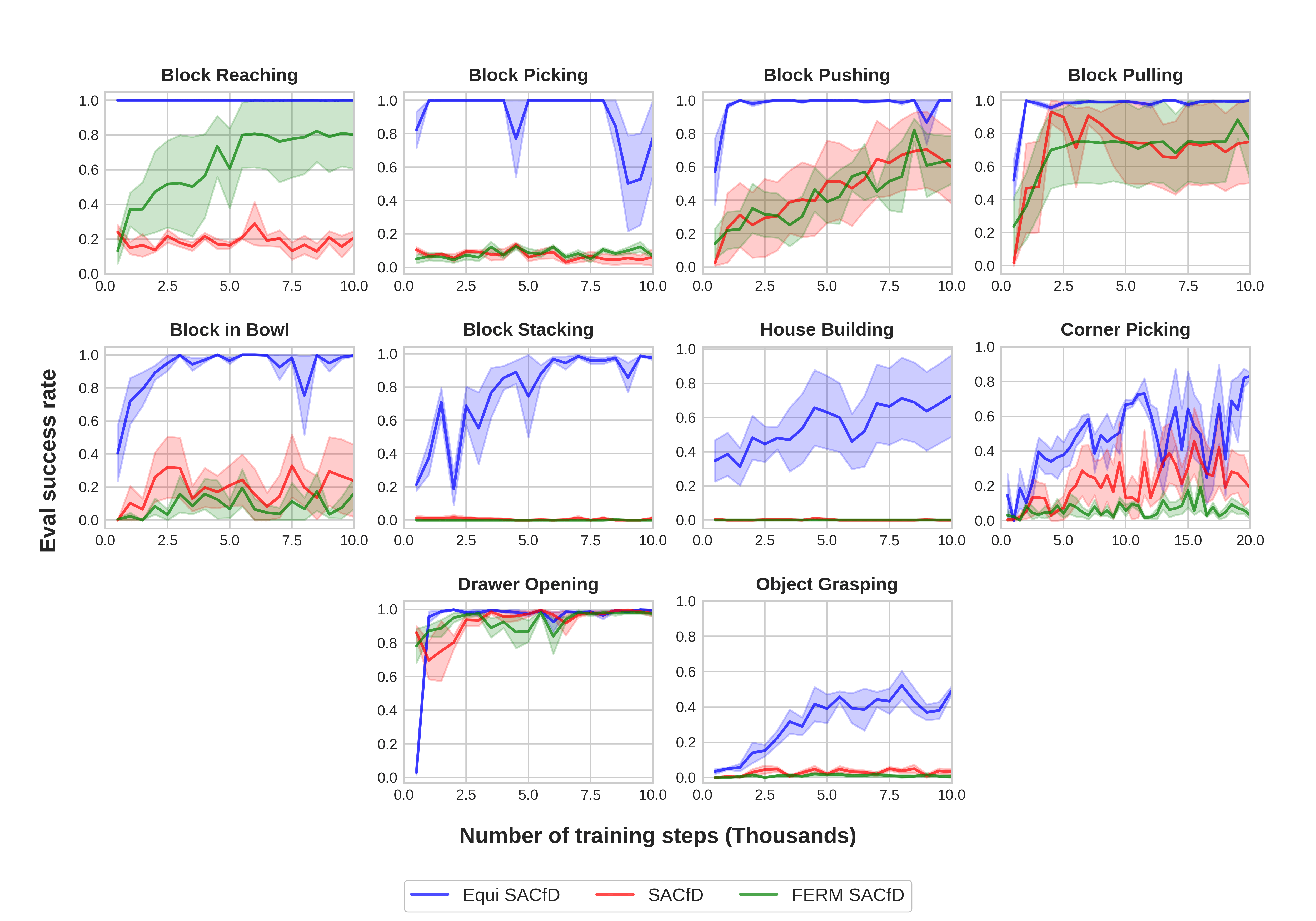}
\centering
\caption{The Close-Loop 3D benchmark result. The plots show the evaluation performance of the greedy policy in terms of the task success rate. The evaluation is performed every 500 training steps. Results are averaged over four runs. Shading denotes standard error.}
\label{fig:close_3d_exp}
\end{figure}

\subsection{Close-Loop 3D Benchmark}
The Close-Loop 3D Benchmark is similar as the Close-Loop 4D Benchmark (Section~\ref{sec:close_ben}), but with the following two changes: first, the environments are initialized with a fixed orientation; second, the action space is $A_\lambda^{xyz} \in \mathbb{R}^4$ instead of $A_\lambda^{xyz\theta} \in \mathbb{R}^5$, i.e., the agent only controls the delta $(x, y, z)$ position of the end-effector and the open-width $\lambda$ of the gripper. 

We provide the same baseline algorithms as in Section~\ref{sec:close_ben}. In this section, we show the performance of SDQfD, Equivariant SDQfD (Equi SDQfD), and FERM SDQfD. Figure~\ref{fig:close_3d_exp} shows the result. Equivariant SACfD (blue) shows the best performance across all tasks. FERM SACfD (green) and SACfD (red) has similar performance, except for Block Reaching, where FERM SACfD outperforms standard SACfD.

\begin{figure}[t]
\includegraphics[width=\linewidth]{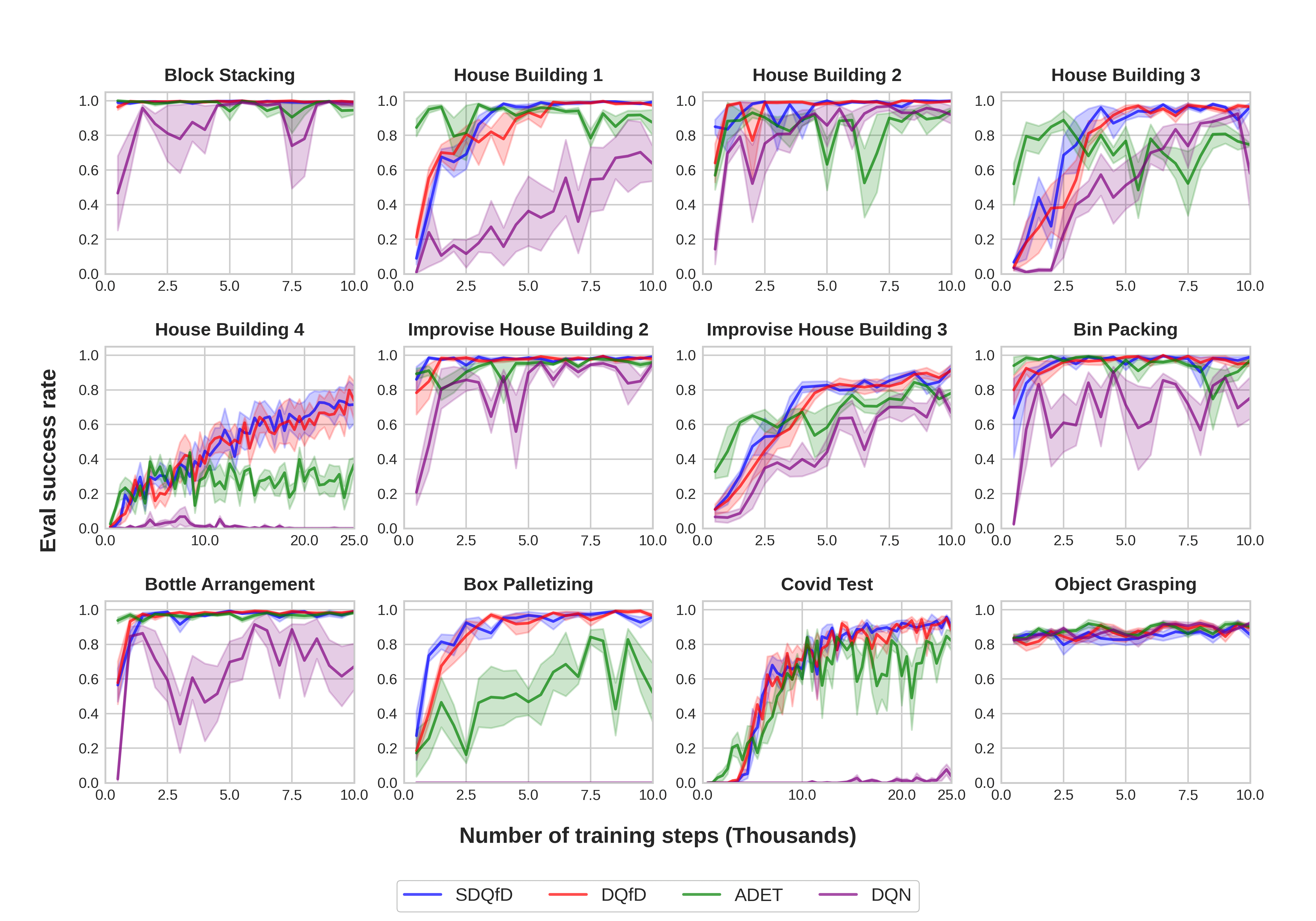}
\centering
\caption{The Open-Loop 3D benchmark result with additional baselines. The plots show the evaluation performance of the greedy policy in terms of the task success rate. The evaluation is performed every 500 training steps. Results are averaged over four runs. Shading denotes standard error.}
\label{fig:open_3d_extra}
\end{figure}

\section{Additional Baselines for Open-Loop 3D Benchmark}
\label{appendix:open_3d_extra}
In this section, we show the performance of three additional baseline algorithms in the Open-Loop 3D Benchmark (Section~\ref{sec:open_ben}): DQfD, ADET, and DQN. We compare them with SDQfD (the algorithm used in Section~\ref{sec:open_ben}). All algorithms are equipped with the Equivariant ASR architecture. Figure~\ref{fig:open_3d_extra} shows the result. Notice that SDQfD and DQfD generally perform the best, while SDQfD has a marginal advantage compared with DQfD. ADET learns faster in some tasks (e.g., House Building 1), but normally converges to a lower performance compared with SDQfD and DQfD. DQN performs the worst across all environments because of the lack of imitation loss.

\begin{figure}[t]
\includegraphics[width=\linewidth]{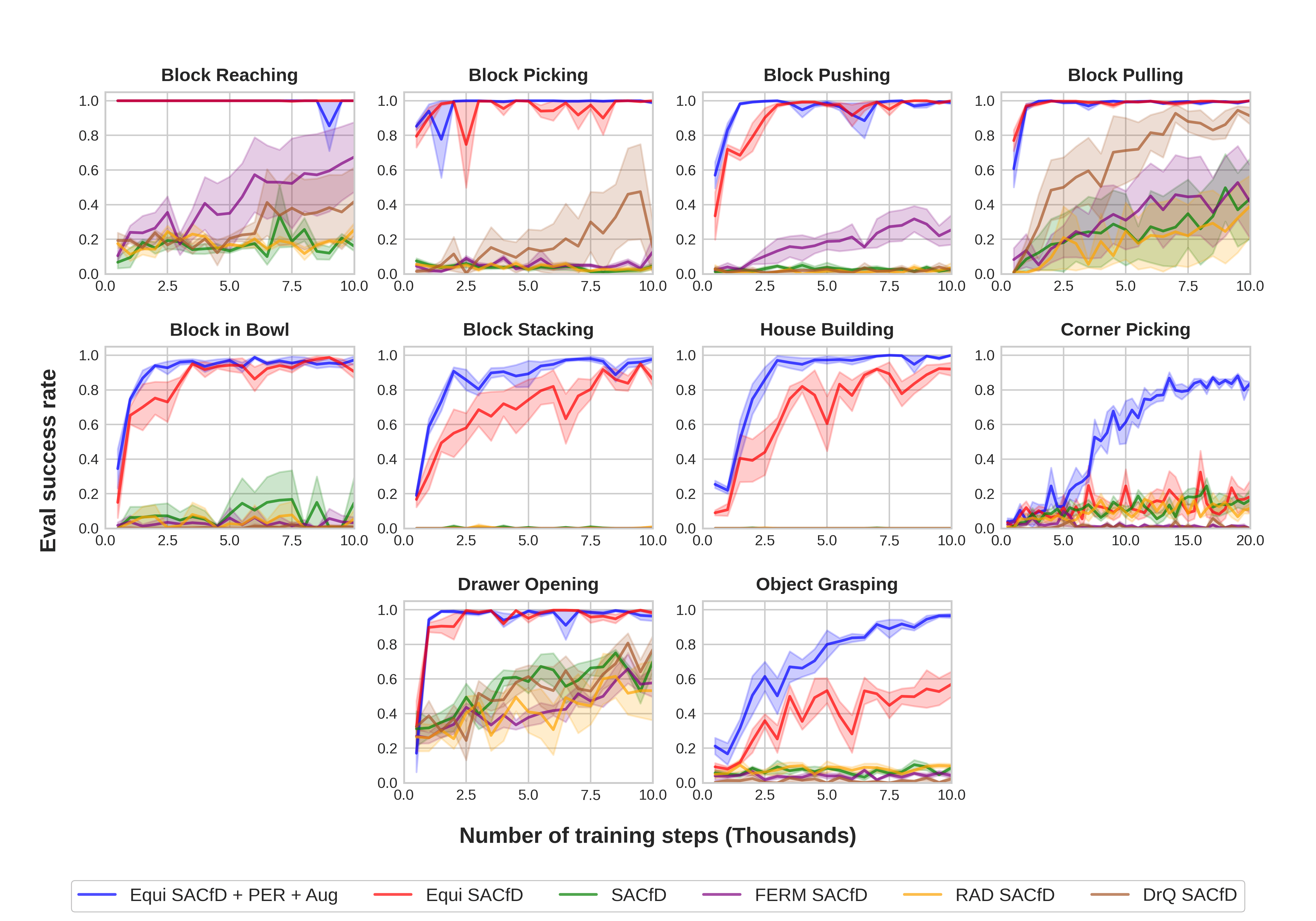}
\centering
\caption{The Close-Loop 4D benchmark result with additional baselines. The plots show the evaluation performance of the greedy policy in terms of the task success rate. The evaluation is performed every 500 training steps. Results are averaged over four runs. Shading denotes standard error.}
\label{fig:close_4d_extra}
\end{figure}

\section{Additional Baselines for Close-Loop 4D Benchmark}
\label{appendix:close_4d_extra}

In this section, we show the performance of two additional baseline algorithms in the Close-Loop 4D Benchmark (Section~\ref{sec:close_ben}): RAD SACfD and DrQ SACfD. As is shown in Figure~\ref{fig:close_4d_extra}, RAD SACfD (yellow) performs poorly in all 10 environments. DrQ SACfD (brown) outperforms FERM SACfD (purple) in Block Picking and Block Pulling, but still underperforms the equivariant methods (blue and red).

\section{Benchmark Details}
\subsection{Open-Loop Benchmark}
In all environments, the kuka arm is used as the manipulator. The workspace has a size of $0.4m\times 0.4m$. The top-down observation $I$ covers the workspace with a size of $128\times 128$ pixels. (In the Rot FCN baseline, $I$'s size is $90\times90$ pixels, and is padded with 0 to $128\times128$ pixels. This is padding required for the Rot FCN baseline because it needs to rotate the image to encode $\theta$.) The size of the in-hand image $H$ is $24\times 24$ pixels for the Open-Loop 2D and Open-Loop 3D benchmarks. In the Open-Loop 6D Benchmark, $H$ is a 3-channel orthographic projection image, with a shape of $3\times 24\times 24$ in the ramp environments, and $3\times 40\times 40$ in the bump environments. We train our models using PyTorch~\cite{pytorch} with the Adam optimizer~\cite{adam} with a learning rate of $10^{-4}$ and weight decay of $10^{-5}$. We use Huber loss~\cite{huberloss} for the TD loss. The discount factor $\gamma$ is $0.95$. The mini-batch size is $16$. The replay buffer has a size of 100,000 transitions. At each training step, the replay buffer will separately draw half of the samples from the expert data and half of the samples from the online transitions. The weight $w$ for the margin loss term of SDQfD is $0.1$, and the margin $l=0.1$. We use the greedy policy as the behavior policy. We use 5 environments running in parallel.

\subsection{Close-Loop Benchmark}
In all environments, the kuka arm is used as the manipulator. The workspace has a size of $0.3m\times 0.3m\times 0.24m$. The pixel size of the top-down depth image $O$ is $128\times 128$ (except for the FERM baseline, where $I$'s size is $142\times 142$ and will be cropped to $128\times 128$). $I$'s FOV is $0.45m\times 0.45m$. We use the Adam optimizer with a learning rate of $10^{-3}$. The entropy temperature $\alpha$ is initialized at $10^{-2}$. The target entropy is -5. The discount factor $\gamma=0.99$. The batch size is 64. The buffer has a capacity of 100,000 transitions. In baselines using the prioritized replay buffer (PER), PER has a prioritized replay exponent of 0.6 and prioritized importance sampling exponent $\beta_0 = 0.4$ as in~\cite{per}. The expert transitions are given a priority bonus of $\epsilon_d=1$. The FERM baseline's contrastive encoder is pretrained for 1.6k steps using the expert data as in~\cite{ferm}. We use 5 environments running in parallel.

\end{document}